\pdfoutput=1
\documentclass[journal]{IEEEtran}

\usepackage{hyperref}
\usepackage{amsmath}
\usepackage{multirow}
\usepackage[english,vlined,linesnumbered]{algorithm2e}
\usepackage[disable]{todonotes}
\usepackage{fullpage}
\usepackage[affil-it]{authblk}
\usepackage{soul,xcolor}
\usepackage[normalem]{ulem}

\usepackage{pdflscape}
\usepackage{fullpage,graphicx}
\usepackage{rotating}

\usepackage{longtable}
\usepackage{supertabular}
\usepackage{multicol}
\hypersetup{hidelinks}

\everymath{\displaystyle}
\DeclareMathSizes{10}{8}{7}{6}   

\newcommand{\gss}{$\gamma$-AM}
\newcommand{\disc}{percentile}
\newcommand{\Disc}{Percentile}
\newcommand{\Sim}{\text{Sim}}
\newcommand{\tfill}{1mm}
\newcommand{\tffill}{-10mm}
\newcommand{\tftfill}{-6mm}
\newcommand{\fffill}{-6mm}
\newcommand{\bfill}{-10mm}
\newcommand{\clusteryea}{8}
\newcommand{\clusterara}{9}
\newcommand{\geneexampleyea}{POP2}
\newcommand{\geneexampleara}{AT2G19450}

\newcommand{\classi}{Rhee \emph{et al.} \cite{rhee2008use}}

\makeatletter
\newcommand{\nosemic}{\renewcommand{\@endalgocfline}{\relax}}
\newcommand{\dosemic}{\renewcommand{\@endalgocfline}{\algocf@endline}}
\let\oldnl\nl
\newcommand{\nonl}{\renewcommand{\nl}{\let\nl\oldnl}}
\makeatother

\newcommand{\colornew}{red}

\newcommand{\dif}{\textbf} 
\newcommand{\diff}{\textcolor{\colornew}}

\renewcommand{\dif}{}
\renewcommand{\diff}{\textbf}

\setstcolor{\colornew}

\begin{document}


\title{Inferring unknown biological function by integration of GO annotations and gene expression data}

\author{G. Leale, A. Bay\'a, D.H. Milone,~\IEEEmembership{Member,~IEEE}, P. Granitto and G. Stegmayer,~\IEEEmembership{Member,~IEEE}.
\IEEEcompsocitemizethanks{%
\IEEEcompsocthanksitem G. Leale is with Center for R\&D of Information Systems (CIDISI), CONICET and Research Institute for Signals, Systems and Computational Intelligence (sinc(\textit{i})), FICH-UNL, CONICET (email: guileale@santafe-conicet.gov.ar).
\IEEEcompsocthanksitem G. Stegmayer and D. H. Milone are with Research Institute for Signals, Systems and Computational Intelligence (sinc(\textit{i})), FICH-UNL, CONICET (email: d.milone@ieee.org).
\IEEEcompsocthanksitem A. Bay\'a and P. Granitto are with CIFASIS, French Argentine International Center for Information and Systems Sciences, UNR-CONICET (email: \{baya,granitto\}@cifasis-conicet.gov.ar).}
\thanks{}
}



\maketitle

\begin{abstract}

Characterizing genes with semantic information is an important process regarding the description of gene products. In spite that complete genomes of many organisms have been already sequenced, the biological functions of all of their genes are still unknown. Since experimentally studying the functions of those genes, one by one, would be unfeasible, new computational methods for gene functions inference are needed. We present here a novel computational approach for inferring biological function for a set of genes with previously unknown function, given a set of genes with well-known information. This approach is based on the premise that genes with similar behaviour should be grouped together. This is known as the \emph{guilt-by-association} principle. 
Thus, it is possible to take advantage of clustering techniques to obtain groups of unknown genes that are co-clustered with genes that have well-known semantic information (GO annotations). Meaningful knowledge to infer unknown semantic information can therefore be provided by these well-known genes.
We provide a method to explore the potential function of new genes according to those currently annotated.
The results obtained indicate that the proposed approach could be a useful and effective tool when used by biologists to guide the inference of biological functions for recently discovered genes. \diff{Our work sets an important landmark in the field of identifying unknown gene functions through clustering, using an external source of biological input.}
A simple web interface to this proposal can be found at http://fich.unl.edu.ar/sinc/webdemo/gamma-am/.

\end{abstract}

\section{Introduction}
Describing genes in terms of biological functions constitutes a major challenge in the study of recently discovered gene products. Considering the large amount of gene data available, it is difficult to perform wet experiments for every condition in each individual gene. Therefore, computational methods are needed as an efficient way to infer biological function for newly discovered genes \cite{pinoli2015computational}. In particular, clustering can be useful to this end, in order to unveil underlying related information within groups of genes.
Clustering is widely used for knowledge discovery, since it is mainly meant to find interesting and previously unknown properties on a given problem. It is commonly divided into three phases: a) measuring the proximity of objects under study. For this phase, studies in bioinformatics make use of the Euclidean distance and the Pearson correlation coefficient, essentially because of their wide availability and ease of use \cite{de2008clustering}. Nevertheless, new metrics are proposed to make better measurements of proximity between objects, for example through the use of path-based dissimilarities, graphs and perceptual organization \cite{nguyen2009novel,baya2011clustering}; b) grouping objects according to this proximity. Literature is profuse regarding this phase \cite{xu2008clustering}, including the development of several clustering algorithms used in bioinformatics \cite{de2008clustering,milone2010omesom,de2013pattern}; and c) evaluating the quality of the formed groups or clusters. There has been a growing interest in this phase over the last years \cite{stegmayer2012biologically,baya2013many, Halkidi01onclustering}.
In the biological field, clustering methods are performed upon the well-known \emph{guilt-by-association} principle \cite{wolfe2005systematic}. This biological assumption implies that genes involved in common biological processes behave similarly \cite{lacroix2008introduction}. Therefore, if a gene with unknown biological function behaves in a similar manner than a well-known one, a strong inference that both genes are involved in the same regulatory process could be made, and thus these genes should be clustered together \cite{usadel2009co}.
\dif{The authors in \cite{wang2004gene} have studied the idea that highly correlated genes have strong similarity also on information coming from the GO ontology. The study has tested the feasibility of applying GO-driven similarity methods for the inference of gene function, concluding that this strategy can lead to better results.}

It can be stated that, in general, information about biological processes is not used in an explicit way in the training patterns when applying clustering algorithms. However, biologists use this knowledge later for validation of clusters and relations found within data \cite{journals/bioinformatics/HandlKK05}. Furthermore, data on well-known biological functions are available and their associations to each data pattern could be made in a straightforward fashion. As a consequence, considering this semantic information explicitly could be very useful while clusters are being formed during training, with the aim of finding results with better quality from a biological point of view. Recently, several new measures have been proposed, in particular on the basis of knowledge representations such as the Gene Ontology (GO) \cite{Consortium2004}. GO is a controlled vocabulary, which considers the \emph{semantics} of each gene as an alternative to traditional experiment-based measures \cite{lord03}. It provides concepts or \emph{terms} organized in a structured set. These terms are systematically used to describe or \emph{annotate} genes. The proximity, or \emph{similarity} between two terms can be thought of as the extent to which they share information in common. This similarity information is contained in the level of specificity of the term that subsumes them, and is named \emph{information content} (IC). As the specificity of the subsumer term raises, the IC is higher, and vice versa \cite{Resnik1999}. Several similarity measures for comparing biological terms based on IC have been developed \cite{Pesquita2009}. Recent studies conclude that the use of a distance that combines both measures based on gene expression data and semantic sources such as ontologies can lead to more biologically meaningful and stable clusters \cite{Kustra2010,Dotan-Cohen2009,milone2014improving}. 
\diff{In this work, we present a novel method that goes a step beyond these cited works, because the proposed approach integrates different sources of data and proposes an original way to measure distance between genes  in order to improve clustering. It also presents an innovative way of inferring GO labels from the better clusters obtained after data fusion and clustering.}

Semantic information such as GO annotations is not always available for all genes within a genome. Therefore, a way of inferring or assigning semantic information to genes that are not systematically described in literature, nor annotated to any GO terms, is needed. 
\dif{Given the vast amount of genomic data, the experimental determination of gene functions is inherently difficult and expensive, therefore automated annotation of gene functions has emerged as a challenging problem in computational biology \cite{radivojac2013large}}.
Although computational methods based on the aforementioned ``guilt-by-association'' principle have been often used \cite{rhee2008use}, most of these methods do not consider all available information, such as for example, biological annotations.
We present here a novel approach, the \emph{Gamma Assignment Method} (\gss{}), which aims to infer the biological function of genes with previously unknown function. The approach has three main steps that include measuring data similarity, taking into account not only expression data but also semantic GO annotations, assigning genes with unknown biological function to clusters formed by genes with well-known semantic information, and characterizing those unknown genes with biological functions through cluster enrichment analysis. 
\diff{An experimental comparison with other methods, on real biological datasets, shows the usefulness and superiority of the proposal.}
\diff{The strength of our approach relies on its simplicity and, at the same time, effective ability to improve the inference results when compared against the classical approach and other related clustering-based methods. Furthermore, an innovative method for proximity measure is introduced for the clustering step, thus providing more meaningful clusters from a biological point of view, before the function inference.}
\diff{Considering these facts, we can state that our proposal moves a step forward in the field of inferring novel GO annotations through the use of available information only from gene expression, and known GO annotations for other genes.}

This work is organized as follows. Section \ref{sec:02} presents a detailed description of the approach. Section \ref{sec:methods} presents the datasets and the performance measures used in this work. Section \ref{sec:results} shows the results obtained, including detailed examples of the application of our approach. The conclusions and future work can be found in Section \ref{sec:conclusions}.

\section{A new approach for inferring biological data function}\label{sec:02}

Our new approach, \gss{}, aims to assign biological function to genes with unknown biological function, given a background gene set with well-known function. This process is performed as follows. 
$A$ is a set of genes with expression data, having also information about the biological function. $B$ is a gene set with expression data and unknown biological function. Both sets are provided as input to \gss{}.
The purpose of our approach is to identify the corresponding function of genes in $B$ by assigning those genes to clusters of genes from $A$ (with well-known function) and inferring the biological function from their co-cluster members.

\begin{figure*}[t!]
\centerline{\includegraphics[scale=0.58]{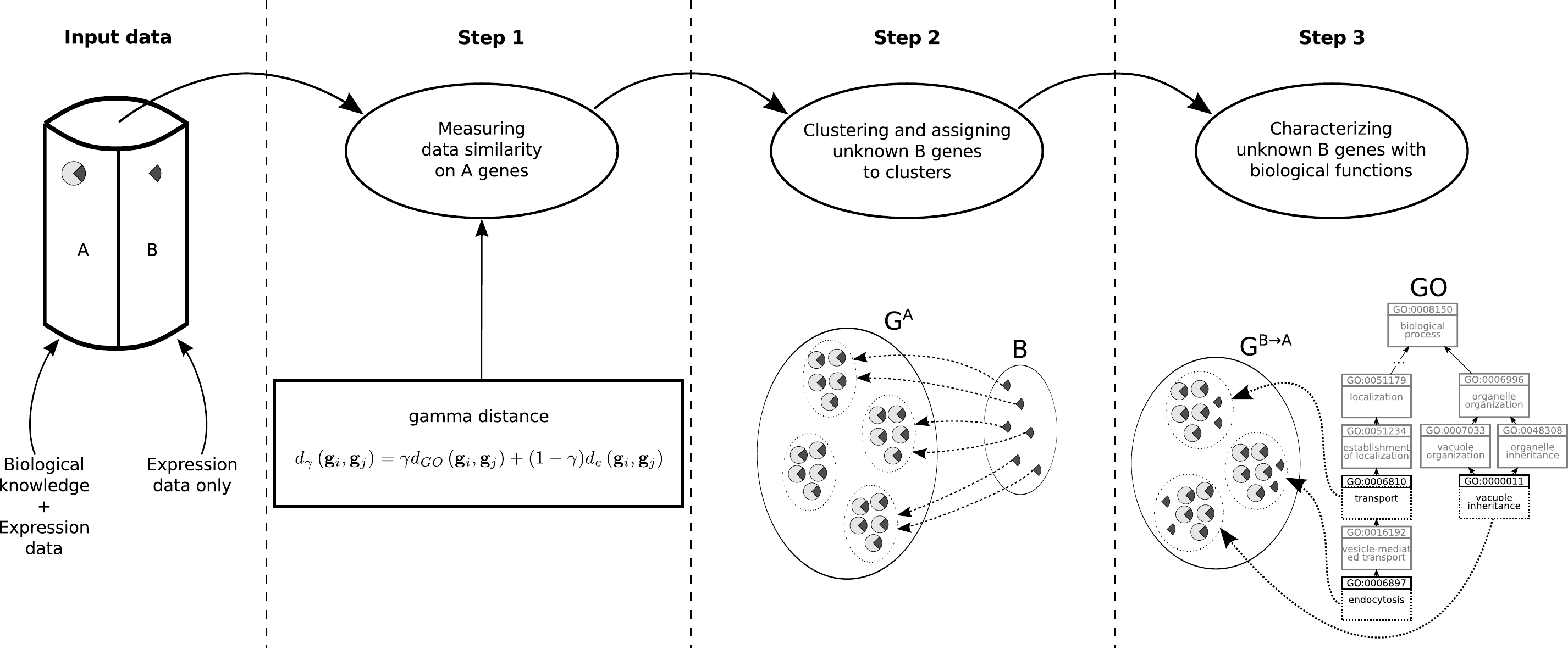}}
\caption{A workflow for the \gss{} approach.}\label{fig:01}
\vspace{-5mm}
\end{figure*}

The proposed approach has three main steps. A workflow describing the whole process is shown in Figure \ref{fig:01}. In Step 1, a distance matrix among genes in set $A$ must be calculated. For this step, a new parameter-based distance, named \emph{gamma distance}, is used in order to combine both available expression and biological knowledge. It should be noted that both types of data in the gamma distance may have different statistical distributions. Consequently, a way of balancing them is needed. Two methods will be proposed to accomplish this issue at this step. In step 2, clusters are formed from set $A$ according to the gamma distance. After that, genes in set $B$ are assigned to the clusters obtained from $A$, forming a partition $G^{B \rightarrow A}$ that contains only the clusters in which the $B$ genes have been assigned. In Step 3, each cluster in $G^{B \rightarrow A}$ is characterized with a set of biological functions. This is achieved through \emph{GO enrichment analysis}, which finds GO terms that are over-represented within a gene subset with respect to a gene background set (in this case, $A$) \cite{Huang2009}. Semantic information for each previously assigned gene from $B$ is then inferred with a set of terms according to the enrichment analysis that has been performed to its corresponding cluster. The following sub-sections explain in detail each step of \gss{}. 

\subsection{Step 1: Measuring data similarity on A genes by combining both expression data and biological knowledge}\label{sec:2_1}
The first step before applying any clustering algorithm on data involves the use of an adequate similarity measure that allows finding highly cohesive and well-separated clusters. Particularly with genes, a natural choice is to group together genes with similar patterns of expression. After that, for clustering, a distance matrix must be built. Each element of the matrix will be a pairwise distance (for example, genes $\mathbf{g}_i$ and $\mathbf{g}_j$) calculated according to $1-\text{similarity}\left(\mathbf{g}_i,\mathbf{g}_j\right)$.
Several distance measures are currently used for clustering biological data in particular \cite{Pesquita2009,eisen1998cluster}. Common choices for measuring the distance between the expression of $\mathbf{g}_i$ and $\mathbf{g}_j$ are the classical Euclidean distance and the Pearson distance. While Euclidean distance can take into account differences in the absolute level of expression of data \cite{Pereira2009}, when the Pearson distance is used positively correlated genes are considered similar (near) to each other and lower and negative correlations indicate not similar (distant) data points \cite{Gibbons2002a}. The standard correlation coefficient might conform appropriately to the intuitive biological notion of what it means for two genes to be ``co-expressed'', because this statistics captures similarity in ``shape'' but places no emphasis on the magnitude of the measurements \cite{eisen1998cluster}. Both distances are usually normalized in the range $\left[0,1\right]$ \cite{Dotan-Cohen2009}. 
\dif{It should be noted that this choice of similarity measure allows the use of any numeric-based gene expression analysis, regardless of the technology used for collecting the data.} 
\dif{Also, multiple expression experiments could be used as well with the proposed method, just by extending each feature vector by adding more dimensions to the gene expression data. In this case, data should be normalized before integration into a single vector. For example, in a simple and straightforward way, each experiment data vector could be divided by its L2 norm, where the square root of the sum of each element, squared, is equal to 1.}

A different kind of measure can be used between pairs of genes considering their relationship in terms of biological knowledge. A common choice of representation for this type of relationship is given through semantic similarity, which can be calculated upon objective biological knowledge representations or \emph{annotations}. Such annotations can be found in ontologies such as GO, where a structured and controlled vocabulary is used to associate biological knowledge to a pre-defined set of descriptions or terms \cite{TheGeneOntologyConsortium2013}. Terms farther from the root describe more specific concepts, whereas terms closer to the root describe high-level abstract concepts. The adoption of ontologies for annotation provides means to compare entities on aspects that otherwise would not be comparable by classical (for example expression-based) distances. 

Semantic similarity measures can be defined as functions that, given two ontology terms or two sets of terms annotating two entities, return a numerical value reflecting the closeness in meaning between them \cite{Pesquita2009}. Several measures have been developed following this approach. In particular, many of them have some issues related to lack of sensitivity with respect to the level of detail within the ontology \cite{Dotan-Cohen2009}. This can lead to some counter-intuitive assumptions, such as similar closeness for two specific and two abstract terms, and equal closeness for every pair of genes which are descendants of the same common ancestor. One measure that solves both aforementioned issues is the Relevance measure \cite{Pesquita2009}, proposed by Schlicker \cite{Schlicker2006} to measure the similarity between terms $t_i$ and $t_j$

\vspace{-4mm}
\begin{footnotesize}
\begin{align}\label{eq:01} 
\Sim{} \left(t_{i},t_{j}\right) &=& \max_{t \in S \left(t_{i},t_{j} \right)} \left\lbrace \frac{2 \log p(t)}{\log p(t_{i})+\log p(t_{j})} \left(1-p(t)\right)\right\rbrace \nonumber \\
&=& \frac{2 I(ms(t_{i},t_{j}))}{I(t_{i})+I(t_{j})}  \left(1-e^{-I(ms(t_{i},t_{j}))}\right),
\end{align}
\vspace{-4mm}
\end{footnotesize}

\noindent where $p(t_i)$ is the probability of finding an instance of a term $t_i$ in GO, computed as the number of genes annotated to $t_i$ or one of its descendants divided by the total number of genes in the ontology; $I\left(t_i\right)$ is the information content of a term $t_i$ that can be quantified as the negative log likelihood of $p(t_i)$; $S \left(t_{i},t_{j} \right)$ is the set of common ancestors between $t_{i}$ and $t_{j}$; $ms\left(t_{i},t_{j}\right)$ is the minimum subsumer or the term $t$ that maximizes $I$ in $S \left(t_{i},t_{j} \right)$. It is the common ancestor between $t_{i}$ and $t_{j}$ with the higher information content, and therefore the closest one to both $t_{i}$ and $t_{j}$, thus $I\left(ms\left(t_{i},t_{j}\right)\right) = \max_{t \in S \left(t_{i},t_{j} \right)} I\left(t\right)$. This measure considers the relative location of the terms with respect to their minimum subsumer, and also the location of the minimum subsumer within the ontology. Minimum subsumers which are very specific will provide higher similarity between the terms subsumed than those located near the root of the ontology. The Relevance measure varies in the range $[0,1]$. 

It is important to consider that this similarity measure is calculated between ontological \emph{terms}, not genes. It should also be noted that a gene is tipically annotated to more than one term. For example, the gene RFC1 from the budding yeast \emph{Saccharomyces cerevisiae} is annotated to all the GO terms shown in Table \ref{tab:table1}. Those terms represent the biological processes in which this gene is involved. Since we want to cluster genes based on a distance measure, a \emph{semantic distance} between genes must be defined first. Let $GO_{\mathbf{g}_i}$ and $GO_{\mathbf{g}_j}$ be the sets of terms annotating the genes ${\mathbf{g}_i}$ and ${\mathbf{g}_j}$. $|GO_{\mathbf{g}_i}|$ and $|GO_{\mathbf{g}_j}|$ are the number of terms in $GO_{\mathbf{g}_i}$ and $GO_{\mathbf{g}_j}$, then

\vspace{-5mm}
\begin{footnotesize}
\begin{align}
d_{GO}(\mathbf{g}_i,\mathbf{g}_j) = 1-\frac{1}{2}\left( \frac{1}{|GO_{\mathbf{g}_i}|} \sum_{\forall t_i \in GO_{\mathbf{g}_i}} \max_{t_j \in GO_{\mathbf{g}_j}} \left\lbrace\Sim{} \left(t_{i},t_{j}\right) \right\rbrace \right.  \nonumber \\
+ \left. \frac{1}{|GO_{\mathbf{g}_j}|} \sum_{t_j \in GO_{\mathbf{g}_j}} \max_{\forall t_i \in GO_{\mathbf{g}_i}} \left\lbrace \Sim{} (t_{i},t_{j}) \right\rbrace \right). \nonumber
\end{align}
\vspace{-4mm}
\end{footnotesize}

This distance is an average of the best pairwise distances from each term in one of the sets to all the terms in the other set.

\subsection*{A new measure for gene clustering: the gamma distance}
Based on the expression and semantic distances described above, a new distance that takes into account both will be used here \cite{leale2013new}. Given the genes ${\mathbf{g}_i}$ and ${\mathbf{g}_j}$, let $d_e\left(\mathbf{g}_i,\mathbf{g}_j\right)$ be one expression distance between them, such as Euclidean or Pearson, and $d_{GO}\left(\mathbf{g}_i,\mathbf{g}_j\right)$ be the semantic distance as defined before. Then, given a set of genes, we propose to use the \emph{gamma distance} between ${\mathbf{g}_i}$ and ${\mathbf{g}_j}$, defined as 

\begin{figure}[t!]
\centerline{\includegraphics[scale=0.20]{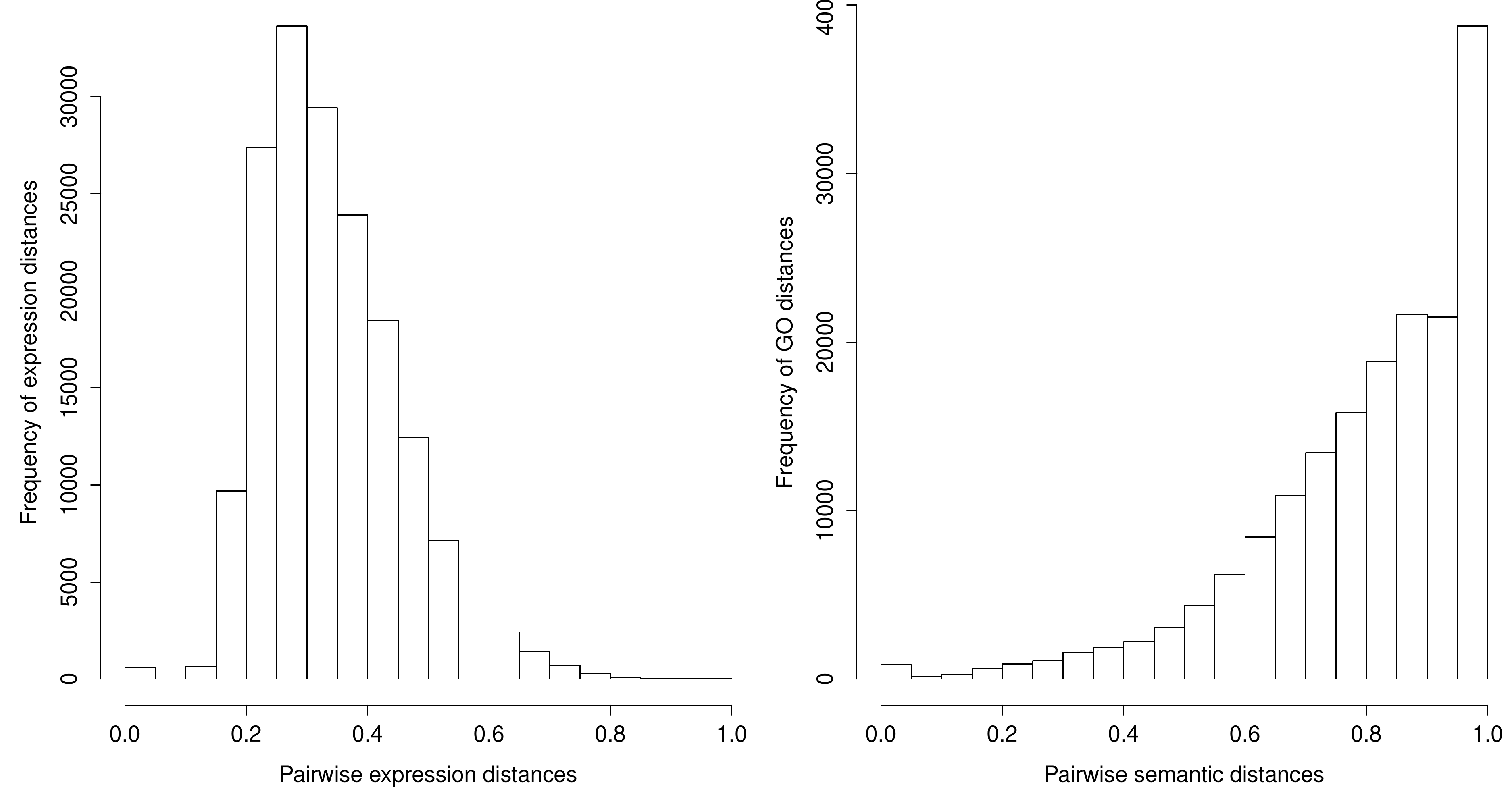}}
\caption{Distance histograms distributions for an example gene dataset (yeast). Expression-based pairwise distances (left) and semantic-based pairwise distances (right).}\label{fig:02}
\vspace{\fffill{}}
\end{figure}

\vspace{-4mm}
\begin{footnotesize}
\begin{align}\label{eq:gamma_distance}
& d_{\gamma}\left(\mathbf{g}_i,\mathbf{g}_j\right)=\gamma d_{GO}\left(\mathbf{g}_i,\mathbf{g}_j\right) + (1-\gamma) d_e\left(\mathbf{g}_i,\mathbf{g}_j\right),
\end{align}
\vspace{-5mm}

\end{footnotesize}
\noindent where the value of the $\gamma$ parameter indicates the importance given to the semantic information in the distance calculation between any pair of genes. A value of $\gamma=0$ corresponds to a pure expression-based distance, and a value of $\gamma=1$ corresponds to a pure semantic-based distance. One important issue regarding the use of this distance is the fact that $d_e$ and $d_{GO}$ come from very different types of data sources. Therefore, it is highly likely for them to present very different statistical distributions. 
This can be a problem in cases where both distance matrices need to be combined as an input to discover gene groups.
For example, Figure \ref{fig:02} shows the histograms distributions of expression-based (left) and semantic-based (right) pairwise distance matrices of genes belonging to an example data set (yeast). The dissimilarity in the distributions in the histograms is very clear. In order to overcome this issue and achieve a more fair influence of both types of distances in the combined gamma distance, the following two methods are proposed.

\begin{table}
\centering
\scriptsize
\begin{tabular}{c|l}
\textbf{GO term }& \textbf{Description }\\ 
\hline 
GO:0007049 & Cell cycle \\ 
GO:0000278 & Mitotic cell cycle \\ 
GO:0051301 & Cell division \\ 
GO:0006281 & DNA repair \\ 
GO:0006298 & Mismatch repair \\ 
GO:0006260 & DNA replication \\ 
GO:0006272 & Leading strand elongation \\ 
\end{tabular} 
\vspace{\tfill{}}
\caption{Illustrative example of GO terms annotating the yeast gene RFC1.}
\label{tab:table1}
\vspace{\tftfill{}}
\end{table}

\subsubsection*{The \disc{} method}
\begin{figure}[t!]
\centerline{\includegraphics[scale=0.20]{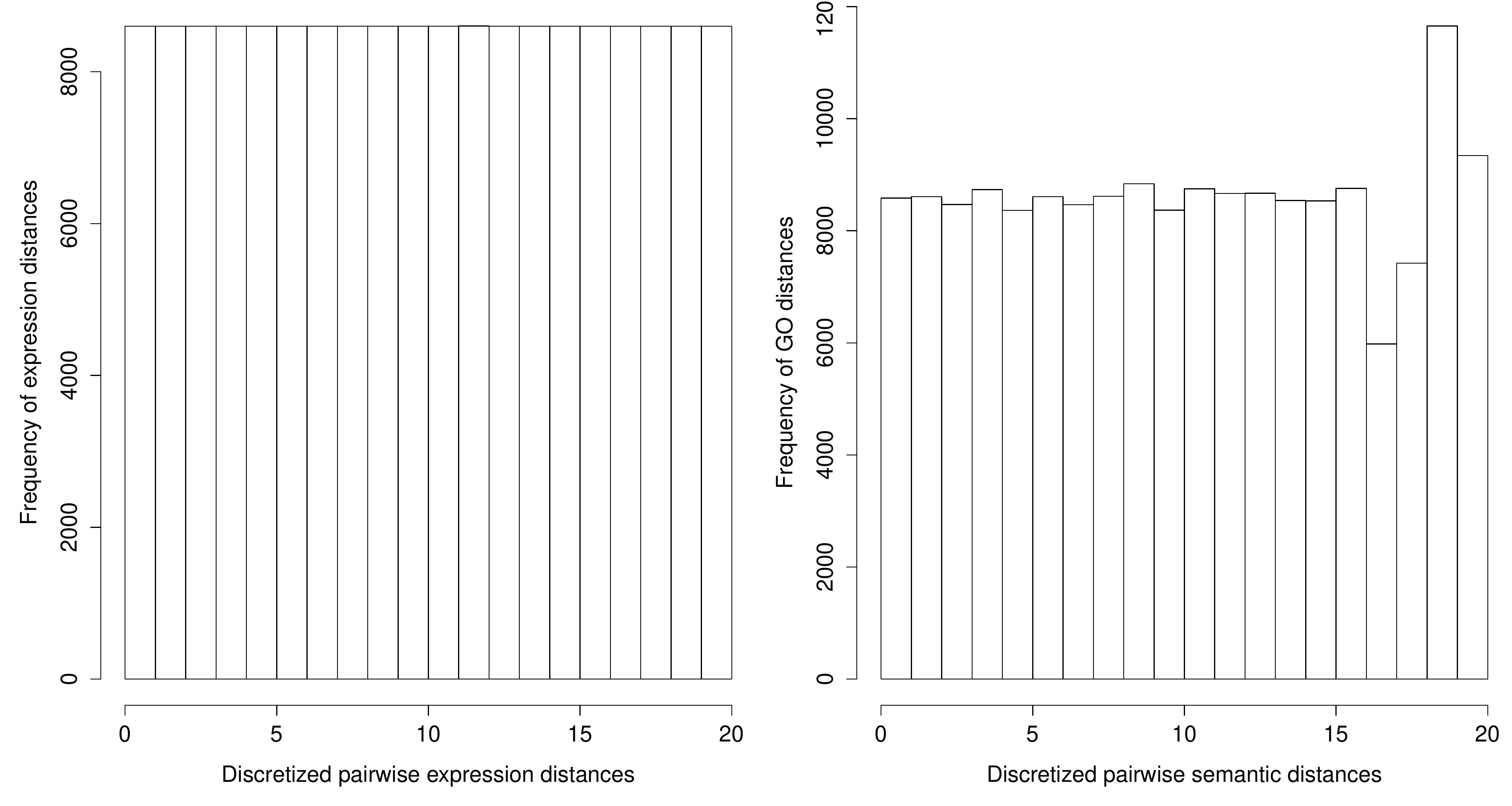}}
\caption{Histograms for the pairwise distances from Figure \ref{fig:02} after applying the \disc{} method.}\label{fig:03}
\vspace{\fffill}
\end{figure}

This method takes into account the original distribution of the pairwise distances corresponding to the gene dataset. The goal is to find a way of homogenizing both semantic and expression distance matrices in order to give them similar importance. The procedure is as follows. Let $D$ be a pairwise distance matrix for the entire gene dataset. The range of possible values for $D$ is obtained and divided into $m$ equal intervals. Then the $i$-th percentile corresponding to each value of $D$ is calculated, and the values are re-distributed into their corresponding $j$-th interval. After this discretization, an approximate uniform distribution is obtained for the values of $D$. The process is applied to both types of distance matrix (expression and semantic). After this process, both distance matrices can be used into the gamma distance with equal weight ($\gamma=0.5$). An example on the application of this method to the distributions of Figure \ref{fig:02} is shown in Figure \ref{fig:03}, where the effect of the method for equalizing the histograms can be clearly seen.

\subsubsection*{The $\gamma$-tuning method}
This method aims to balance the weight given to the original pairwise distance matrices used within the gamma distance by automatically finding an appropriate value for the parameter $\gamma$. The detailed procedure is as follows. First of all, a random subsampling is made out of the original set $A$ of gene measures (both expression and semantic data), generating two subsets: $A_1$ and $A_2$. Semantic-based data from $A_2$ is removed. After that, a clustering algorithm is applied only on $A_1$ with a fixed value of $\gamma$ and $k$, obtaining a clustering partition, $G^{A_1}$. Then, each gene from $A_2$ is assigned to one cluster $\Omega^{A_1}_j / \Omega^{A_1}_j \in G^{A_1}$, $j=1,\ldots,k$ according to the minimum expression-based distance between the gene expression data and the centroid of each cluster. After all genes in $A_2$ have been assigned to a cluster $\Omega^{A_1}_j$, a new partition $G^{A_2 \rightarrow A_1}$ is defined as the one having only those clusters $\Omega^{A_1}_j$ which have genes from $A_2$. Semantic-based information is then restored to genes in $A_2$ and a global measure of compactness is calculated for $G^{A_2 \rightarrow A_1}$. This is performed as follows: the minimum semantic distance between each assigned gene from $A_2$ and genes from $A_1$ clustered together is calculated and averaged through a number $n$ of runs.  With this measure, it is possible to obtain an automatic estimation of the biological quality of the final clusters obtained for a specific value of $\gamma$. 
These $n$ runs are performed for each value of $\gamma$ in an interval of possible $\gamma$ values ranging from $0$ to $1$, with steps of $0.05$. The best value for $\gamma$ will be selected as the one where the global compactness of the solution $G^{A_2 \rightarrow A_1}$ is minimum.

\begin{figure}[t!]
\centerline{\includegraphics[scale=0.23]{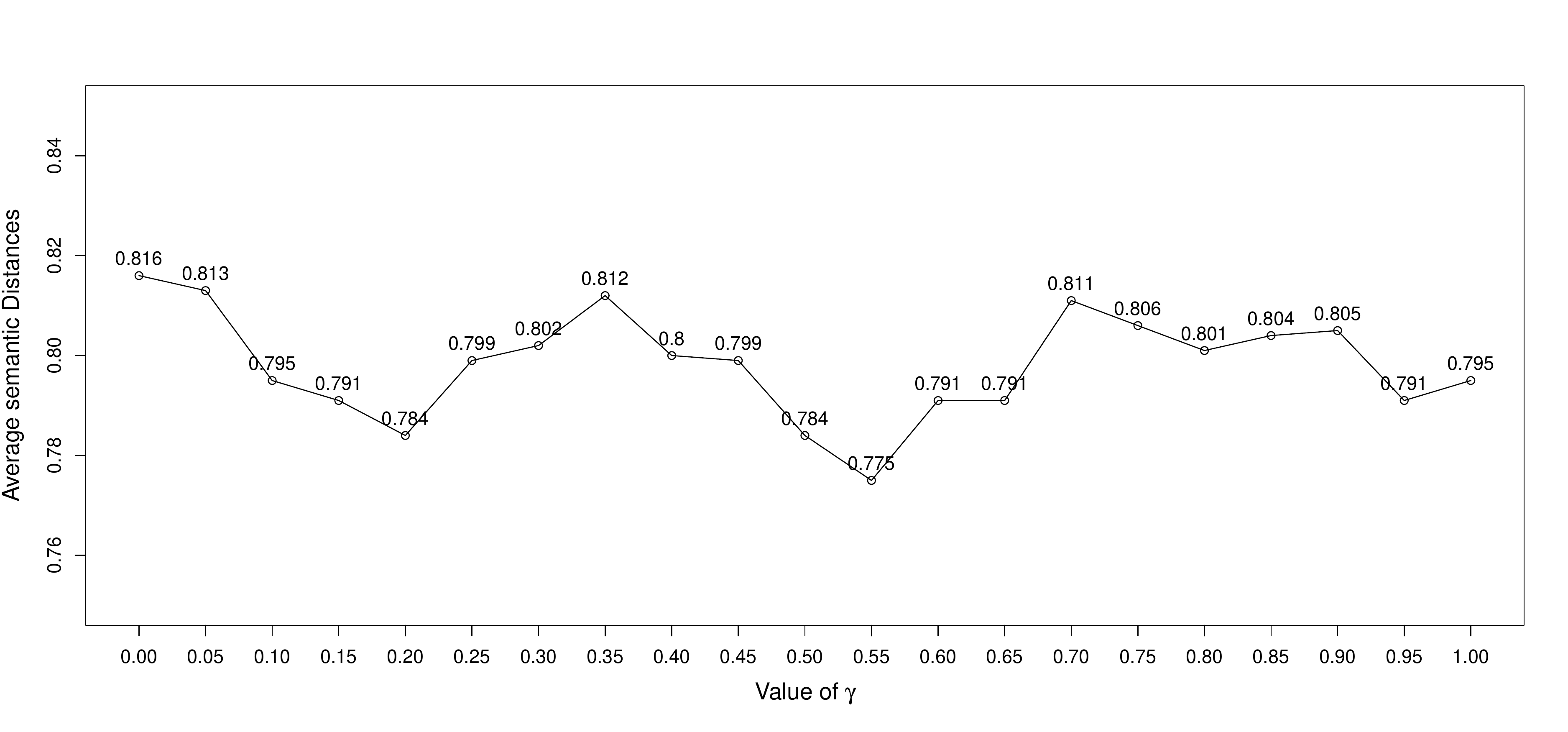}}
\caption{ Example of application of the $\gamma$-tuning method on the yeast data set.}\label{fig:04}
\vspace{\fffill{}}
\end{figure}

An example of a plot with values obtained by using the proposed $\gamma$-tuning algorithm on the yeast dataset is shown in Figure \ref{fig:04}. Each point represents the compactness through all the $n$ runs performed for each $\gamma$, thus indicating the closeness between genes regarding GO annotations. The lowest value (indicating the best semantic compactness of the clusters) will be used as the most appropriate $\gamma$ for the gamma distance in the next step. From the figure, it can be clearly seen that the most appropriate value of $\gamma$ to be used for this data set is $0.55$, corresponding to a minimum average semantic distance of $0.775$.

\subsection{Step 2: Clustering and assigning genes in B.}\label{sec:2_2}

The gamma distance can be used as input to a clustering algorithm in order to obtain more biologically relevant groups of genes. 
\diff{It is important to note that an adequate choice of data representation is crucial for obtaining well-formed and meaningful clusters. Thus, the use of this integrated distance measure, which takes into account domain-specific information as well as experimental data, is key for the achievement of this important goal.}
It must be remembered that $d_\gamma$ is a combined distance between expression and biological data, having a GO-based part which summarizes semantic information from several GO annotations based on similarities rather than on interval-scaled measurement values \cite{Kaufman2005}. Pairwise expression distances between genes and centroids can be calculated, but a distance between a GO-based data point and an artificial centroid cannot be obtained. Thus, a gene must be used as the representative object for each cluster.

\begin{algorithm}[t] \label{alg:algorithm-main}
\footnotesize 
 \SetAlgoLined
 \DontPrintSemicolon
\nonl \hrulefill \\
\nonl {\small \textbf{Algorithm 1:} Clustering and assignment of genes with unknown function to gene clusters with well-known function}\\

\nonl \hrulefill \\
 \KwIn{\\
 
$\quad$ $A$: Set of genes having both expression-based data and GO-based data \\
$\quad$ $B$: Set of genes having expression-based data only \\
$\quad$ $k$: Number of clusters. \\
$\quad$ $\gamma^*$: Weight of GO-based information.
  }
 \KwOut{\\
 
$\quad$  $G^{B \rightarrow A}$: Partition of $B$ assigned to clusters from $A$
	 
 }
  \Begin{

			Set the initial gene-centroid $\mathbf{gc}_1$ as the gene $\mathbf{g}_i$ that minimizes the gamma distance to all other genes in $A$:\\ \label{alg:initial-gc}
			\nonl $\mathbf{gc}_1 \leftarrow \arg\min_{\forall \mathbf{g}_i} \sum_{j \neq i}{d_{\gamma^*}(\mathbf{g}_i,\mathbf{g}_j)}, \quad \mathbf{g}_i, \mathbf{g}_j \in A$\\

			$\Gamma \leftarrow \left\lbrace\mathbf{gc}_1\right\rbrace$ \label{alg:assign-Gamma}\\

			$n \leftarrow 2$ \\
					
			\While{$n \leq k$}{					
						
				\For{each gene $\mathbf{g}_i \in A - \Gamma$}{
				
					Calculate candidate score: \label{alg:sum-scores}
					\nonl $S_{\mathbf{g}_i} = \sum_{\substack{\mathbf{g}_j \not \in \Gamma \\ \mathbf{g}_j \neq \mathbf{g}_i}} \left( d_{\gamma^*}(\mathbf{g}_j,\mathbf{g}_i)-\min_{\forall \mathbf{gc} \in \Gamma}\left\lbrace d_{\gamma^*}(\mathbf{g}_j,\mathbf{gc}) \right\rbrace \right)$
						
				} 
			
				$\mathbf{gc}_n \leftarrow \arg\max_{\forall \mathbf{g}_i} S_{\mathbf{g}_i}$  \label{alg:select-new-gc}		
			
				$\Gamma \leftarrow \left\lbrace\mathbf{gc}_n\right\rbrace$  \label{alg:assign-gc-to-Gamma}\\
				
				$n \leftarrow n+1$\\
			
			}
			
			Assign each $\mathbf{g}_j \in A - \Gamma$ to a cluster $\Omega^A_n$ according to:
			$\mathbf{g}_j \in \Omega^A_n \Leftrightarrow d_{\gamma^*}(\mathbf{g}_j,\mathbf{gc}_n) < d_{\gamma^*}(\mathbf{g}_j,\mathbf{gc}_m) \quad \forall n \neq m$ \label{alg:assign-to-gc} 

			Conform a partition $ G^{A} \leftarrow \left\lbrace \Omega^A_n \right\rbrace$ \label{alg:conform-g-a}

			\Repeat{no new replacements can be made \label{alg:until}}{
			
				\For{each gene-centroid $\mathbf{gc} \in \Gamma$}{
					Calculate: $R \leftarrow \sum_{\mathbf{g}_j \not \in \Gamma} \min_{\forall \mathbf{g}_i \in \Gamma}\left\lbrace d_{\gamma^*}\left(\mathbf{g}_j, \mathbf{g}_i \right)\right\rbrace $\\
					\For{each gene $\mathbf{g}_j \not\in \Gamma$}{
			
						Replace $\mathbf{gc}$ by $\mathbf{g}_j$: $\Gamma' \leftarrow \Gamma - \left\lbrace \mathbf{gc} \right\rbrace \cup \left\lbrace \mathbf{g}_j \right\rbrace$
						\label{alg:replace} \\
						
						Calculate: $R' \leftarrow \sum_{\mathbf{g}_j \not \in \Gamma'} \min_{\forall \mathbf{g}_i \in \Gamma'}\left\lbrace d_{\gamma^*}\left(\mathbf{g}_j, \mathbf{g}_i \right)\right\rbrace $  \label{alg:calculate-R} \\
						\lIf{$R'<R$}{$\Gamma \leftarrow \Gamma'$ \label{alg:maintain}} 
													
					}
					
				}
			
			}
					
			Assign each $\mathbf{g}_h \in B$ to a cluster $\Omega^{B \rightarrow A}_n$ according to:
			$\mathbf{g}_h \in \Omega^{B \rightarrow A}_n \Leftrightarrow d_e (\mathbf{g}_h,\mathbf{gc}_n) < d_e(\mathbf{g}_h,\mathbf{gc}_m) \quad  \forall n \neq m$ \label{alg:assign-B}
			
	Conform a partition $ G^{B \rightarrow A} \leftarrow \left\lbrace \Omega^{B \rightarrow A}_n \right\rbrace$ \label{alg:conform-g-b-a}
	
	}

\nonl \hrulefill \\
\vspace{-7mm}
\end{algorithm}

Taking these facts into account, we propose the Algorithm \ref{alg:algorithm-main}. Let $A$ be a set of genes having both expression and semantic data. Let $B$ be a set of genes having only expression data, $k$ is the number of gene clusters, and $\gamma^*$ is an appropriate weight for the GO-based information used in the gamma distance. This value is determined by one of the methods mentioned in Section \ref{sec:2_1}.

At first, an initial \emph{gene-centroid} $\mathbf{gc}_1$ is defined as the gene that minimizes the gamma distance to all other genes in A (line \ref{alg:assign-Gamma}).
After that, an iterative process is followed until $k$ gene centroids are found.
Let $\mathbf{g}_i \in A$, $\mathbf{g}_i \neq \mathbf{gc}_1$ be a gene from the dataset, which is a candidate to become a new gene-centroid.
For each $\mathbf{g}_i$, a \emph{candidate score} is calculated as $S_{\mathbf{g}_i}$, which takes into account, for each remaining gene $\mathbf{g}_j$ in the dataset, its gamma distance to $\mathbf{g}_i$ and also to its closest gene-centroid (line \ref{alg:sum-scores}), in order to determine if $\mathbf{g}_i$ must be a new gene centroid. 
The gene $\mathbf{g}_i = \arg \max_{\forall \mathbf{g}_i} S_{\mathbf{g}_i}$ will be chosen as a new gene-centroid (line \ref{alg:select-new-gc}) and included into $\Gamma$ (line \ref{alg:assign-gc-to-Gamma}), since the highest candidate score indicates that $\mathbf{g}_i$ is closer to more genes than any other existing gene-centroid. 
After $k$ gene-centroids have been determined, all the rest of the genes from the $A$ set are assigned to one of the $k$ clusters $\Omega^A_n$ (with gene-centroids $\mathbf{gc}_n$) according to its minimum distance to the centroid (line \ref{alg:assign-to-gc}).
Finally, the gene-centroid set $\Gamma$ can be further refined, as suggested in \cite{Kaufman2005}. Each gene-centroid $\mathbf{gc}$ is replaced by another object $\mathbf{g}_j$ which is not a gene-centroid, conforming a new set $\Gamma'$ (line \ref{alg:replace}), and the sum of the pairwise gamma distances from all the other genes to their closest gene-centroids, $R'$, is calculated (line \ref{alg:calculate-R}). If this value is lower than the one obtained with the original gene-centroid set $\Gamma$, the partition is improved, and thus the new gene-centroid configuration remains for the clusters (line \ref{alg:maintain}). This process is repeated until there are no new replacements to be made for the gene-centroids (line \ref{alg:until}). Once clusters $\Omega^A_n$ with all genes from $A$ have been set, forming the partition $G^{A}$, each gene $\mathbf{g}_h \in B$ is assigned to one cluster $\Omega^A_n$ according to the minimum expression-based distance to its corresponding gene-centroid $\mathbf{gc}_n$. These clusters with assigned $B$ genes are named $\Omega^{B \rightarrow A}_n$ (line \ref{alg:assign-B}). A new partition named $G^{B\rightarrow A}$ is then formed, which contains only the clusters from $G^{A}$ which have assigned genes from $B$ (line \ref{alg:conform-g-b-a}).
The biological knowledge corresponding to the genes in $B$ will be inferred according to the enrichment of the clusters where $B$ genes have been included.


\begin{table*}[t!]
\centering
\scriptsize
\begin{tabular}{l|l|l|l}

\textbf{Cluster genes} & \textbf{GO terms} & \textbf{GO labels} & \textbf{GO location} \\ \hline
MSR1 & Gene expression & GO:0010467 & \multirow{3}{*}{\includegraphics[height=70mm,trim=0 0 0 0, clip]{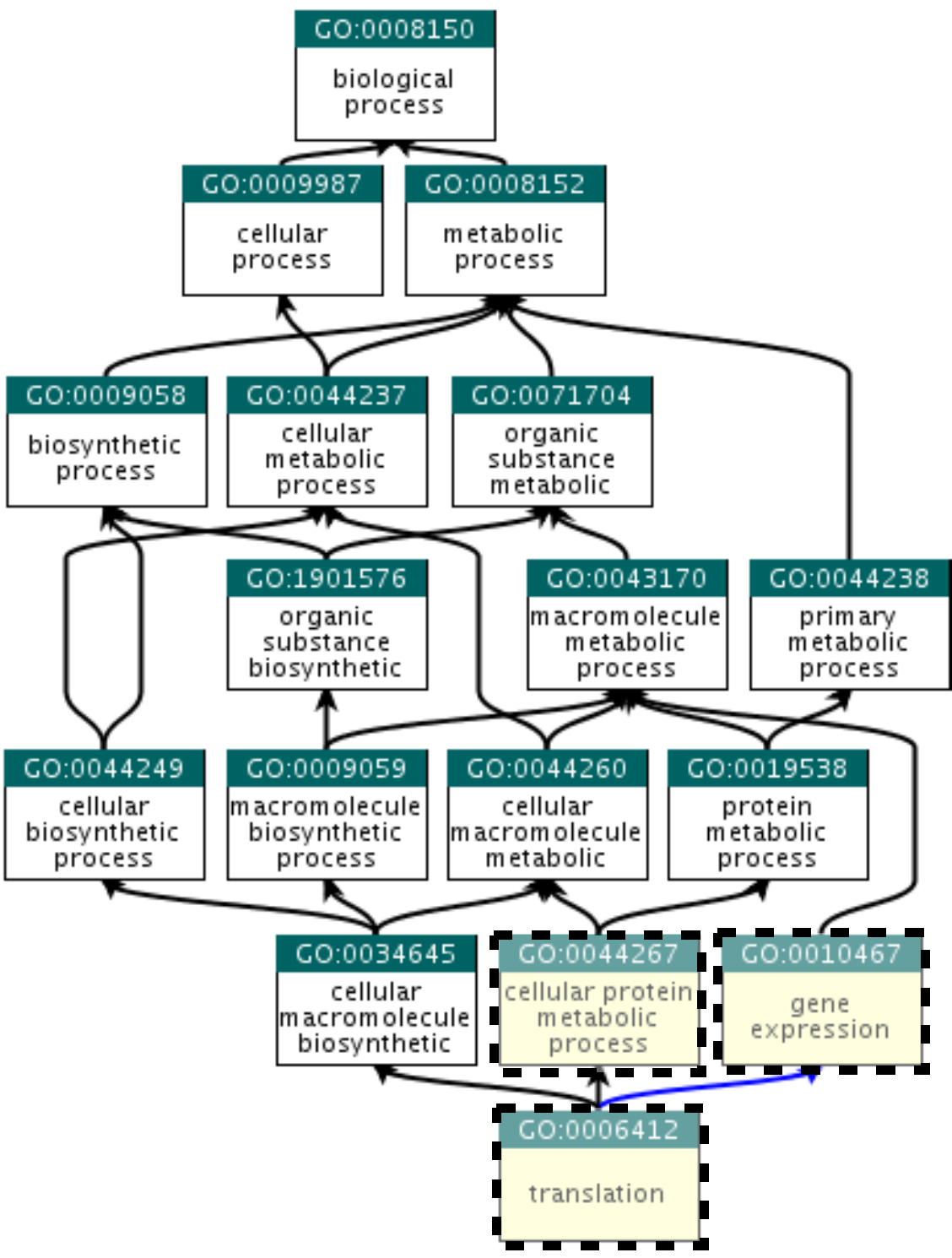}} \\ 
MRPL8 &   &  &  \\ 
RPL17A &  & &  \\ 
SUI1 &  & &  \\ 
ALA1 &  &  &  \\ 
HCA4 &   & &  \\
SQT1 &  &  &  \\ 
NOP56 & & &  \\ 
GAR1 &  &  &  \\ 
NSR1 &  &  &  \\ 
HMT1 &  &  &  \\ 
RPS18A &  &  &  \\ 
RPS14B & Cellular protein  & GO:0044267 &  \\ 
RPL37B &  metabolic process &  &  \\ 
RPL40A &  &  &  \\ 
\ldots &  &   &  \\ 
\ldots &  &   &  \\  
\ldots &  &   &  \\  
\ldots &  &   &  \\  
\ldots &  &   &  \\  
\ldots &  &   &  \\  
\ldots &  &   &  \\  
\ldots &  &   &  \\  
\ldots &  &   &  \\  
\ldots &  & &  \\  
\textbf{(unknown gene)} & Translation  & GO:0006412 &  \\ 
 \end{tabular}
 \vspace{\tfill{}}
\caption{Example of enrichment analysis for a cluster of yeast genes from A with well-known biological function. A gene from B with unknown function is shown in bold.}
\label{tab:table2}
\vspace{\tffill{}}
\end{table*}

\subsection{Step 3: Characterizing unknown B genes with biological functions through cluster enrichment analysis}\label{sec:2_3}
For each cluster in $G^{B\rightarrow A}$, a representative set of biological functions is obtained through the use of enrichment analysis \cite{Huang2009}. This method identifies the functional terms (in this case, GO annotations) which are over-represented in a gene subset with respect to a background gene set. Those terms comprise the biological knowledge associated to the clusters \cite{Hung2012}. The aim of enrichment analysis is to obtain a subset of GO terms for each cluster from $G^{B\rightarrow A}$, using the genes with well-known information as input and the complete gene set $A$ as background. A list of GO terms ordered by $p$-value is provided as result. These terms will characterize the genes from B that were assigned to each cluster. 
\dif{Although annotating a gene to a specific GO term implies its annotation to all its parent terms, in this work we only consider the specific GO terms without its parents for the characterization, in order to make more accurate inferences.}

An example with genes from yeast is shown in Table \ref{tab:table2}. Genes in the first column belong to one cluster from $A$ that has been assigned a gene from $B$ (let us suppose an unknown gene, shown in bold) as described above. The procedure calculates the enrichment $p$-value, obtained as the number of genes in a set that match a given biological function as compared to pure random chance. This calculation can be performed with the aid of well-known statistical methods \cite{khatri2005ontological}. A variety of tools have been developed to apply enrichment analysis \cite{wolfe2005systematic,Huang2009}. Biological functions passing a pre-defined enrichment $p$-value threshold are considered interesting for the input set and returned in an ordered fashion with respect to that value \cite{Huang2009}. In the example, the resulting three biological functions (Gene expression, Cellular protein metabolic process, and Translation) are shown with their GO labels in the second and third column, respectively. The last column shows the location of the GO terms within the GO-Biological Process ontology, enclosed in a dashed line. These three terms resulting from the enrichment analysis of the cluster are used to characterize the unknown gene from B. Finally, this procedure is performed for all clusters from the partition $G^{B \rightarrow A}$ in order to effectively assign biological function to all genes with unknown information from set $B$.

\section{Materials and performance measures} \label{sec:methods}
This section presents the datasets and validation measures used in the experiments.
\subsection{Datasets}
\subsubsection*{YEAST dataset}
This dataset consists of gene expression data from the budding yeast \emph{Saccharomyces cerevisiae}. Several characteristics such as diauxic shift, mitotic cell division cycle and sporulation, were collected in order to study cluster analysis of expression patterns. The activities of collecting and preprocessing the dataset are thoroughly explained in \cite{eisen1998cluster}. From an original dataset of 2467 genes, only those with no missing values were considered. A filtering process to include only those genes annotated to the GO \emph{Biological Process} category \dif{and to exclude those with evidence code ``ND'' (no biological data available) was also applied}. The final dataset has 587 genes.
\dif{For these genes, there are 79 microarray-expression values. Regarding annotation data, there are 1054 unique GO terms for a total of 2845 annotations in this dataset.}

\subsubsection*{ARA dataset}
This dataset comprises genes measured in \textit{Arabidopsis thaliana} leaves. The original work was aimed to study the effects of cold temperatures on circadian-regulated genes in this plant \cite{Espinoza2010}. Genes under light-dark cycles at two control temperatures (20${}^\circ$C and 4${}^\circ$C) and also involved in diurnal cycle and cold-stress responses were selected for the study. From a total of 1549 genes only those annotated to the \emph{Biological Process} category of the Gene Ontology were considered. Genes annotated into GO but \dif{marked with evidence code} ``ND'' were removed. The final dataset used here has 1042 genes.
\dif{For these genes, there are 32 microarray-expression values. In terms of annotation data, 918 unique GO terms were considered for a total of 4470 annotations in this dataset.}

\subsection{Performance measures} 
In this subsection, the following notation is used: the gene dataset is formed by $\mathbf{g}_i$ data samples; $\Omega_c$ is the set of samples that have been grouped in the cluster $c$; $|\Omega_c|$ is cluster size; $k$ is number of clusters.

\subsubsection*{Semantic Compactness} 
A measure to assess the quality of the biological assignment procedure has been defined as follows. Here we will denominate $\mathbf{g}_{i} \in \Omega^{A_2}_c$ and $\mathbf{g}_{j} \in \Omega^{A_1}_c$ the genes clustered in $\Omega_c = \Omega^{A_2}_c \cup \Omega^{A_1}_c$ that, prior to the assignment, belonged to subsets $A_2$ and $A_1$, respectively. Once the partition $G^{A_2 \rightarrow A_1}$ has been obtained, for each gene $\mathbf{g}_{i} \in \Omega^{A_2}_c$, semantic distances to all genes $\mathbf{g}_{j} \in \Omega^{A_1}_c$ clustered together with it in the same cluster $\Omega_c$ are calculated. The minimum semantic distance among them is then selected. 
Therefore, $SC_i$ for a gene $\mathbf{g}_{i} \in \Omega^{A_2}_c$ is defined as
\begin{equation}
 SC_i = \min_{\mathbf{g}_i \in \Omega^{A_2}_c ; \mathbf{g}_j \in \Omega^{A_1}_c  } d_{GO}(\mathbf{g}_{i},\mathbf{g}_{j}).
 \label{eq_bc}
 \end{equation}

\noindent Finally, the overall $SC$ for the whole resulting partition $G^{A_2 \rightarrow A_1}$ is calculated as the average of $SC_i$ for all genes $\mathbf{g}_{i} \in \Omega^{A_2}_c$.
Since $SC$ is based on semantic distances, a lower value represents better compactness, thus better quality. 


\subsubsection*{Biological homogeneity index}
The Biological Homogeneity Index ($BHI$) measures the quality of the clusters on a biological basis. It can be thought of as an average proportion of gene pairs with matched GO terms clustered together \cite{datta2006methods}. Let $F\left(GO_{\mathbf{g}_{i}},GO_{\mathbf{g}_{j}}\right)$ be an indicator function that has the value 1 if $\mathbf{g}_{i}$ and $\mathbf{g}_{j}$ are annotated with at least one term in common, and 0 in any other case. Then
\begin{equation}
 BHI = \frac{1}{k} \sum_{j} \frac{1}{|\Omega_c|\left(|\Omega_c|-1\right)} \sum_{ \mathbf{g}_i \neq \mathbf{g}_j \in \Omega_c} F\left(GO_{\mathbf{g}_{i}},GO_{\mathbf{g}_{j}}\right).
 \label{eq_bhi}
 \end{equation}
$BHI$ can be interpreted as the proportion of common GO annotations within the obtained clusters and it varies in the range $[0,1]$. A value of $BHI$ close to 1 indicates that the clusters are more homogeneous in terms of biological meaning.
\dif{It must be noted that, as mentioned in the Subsection \ref{sec:2_3}, only specific GO terms were taken into account. Thus, the indicator function $F\left(GO_{\mathbf{g}_{i}},GO_{\mathbf{g}_{j}}\right)$ will have the value 1 only when both terms are annotated to at least one GO specific term. As a consequence, there may be many cases when the indicator function has a value of 0, as there are no specific GO matching terms for $\mathbf{g}_{i}$ and $\mathbf{g}_{j}$. This may lead to low values for the overall BHI, although not strictly zero. In spite of this issue, these low BHI values can still be used to obtain a good estimate of the biological quality of the clustering results}.

\subsubsection*{Biological compactness} 
A new measure to evaluate the biological quality of the final gene clustering partition is defined here. Biological Compactness ($BC$) measures the average of the pairwise semantic distances among all elements in each cluster. Thus biological compactness is defined for the cluster $\Omega_c$ as
\begin{equation}
 BC_c = \frac{1}{|\Omega_c|} \sum_{\mathbf{g}_i \in \Omega_c} \sum_{\mathbf{g}_j \in \Omega_c}  d_{GO}(\mathbf{g}_i,\mathbf{g}_j).
 \label{eq_bc}
 \end{equation}
A low value of $BC$ means a cluster with close elements in terms of semantic distances, which can be interpreted as a higher amount of GO-based information in common within each cluster. The overall biological compactness for a solution can be calculated as $BC = \frac{1}{k} \sum_c BC_c$. Values of $BC$ closer to 0 indicate that the clusters are better from a semantic point of view.

\subsubsection*{Fowlkes-Mallows index} 
The Fowlkes-Mallows index is a measure used for external validation of clustering results \cite{fowlkes1983method}. This index is applied here to evaluate the quality of the assignment procedure performed on B genes. Consider a dataset with $N$ elements. Given two clustering solutions for the dataset with $k$ clusters, $C$ and $C'$, FM is defined as 

\begin{equation}
B_k = \frac{T_k}{\sqrt{P_k Q_k}} 
\end{equation}

\noindent where $T_k = \sum_i \sum_j m_{ij}^2-N$, $P_k = \sum_i \left(\sum_j m_{ij}\right)^2 - N$ and $Q_k = \sum_j \left(\sum_i m_{ij}\right)^2 - N$, and $m_{ij}$ is an element from a contingency matrix $M$ obtained between $C$ and $C'$. $P_k$  and $Q_k$ can be interpreted as the probabilities of obtaining a random pair of patterns belonging to the same cluster in $C$ and $C'$, respectively. Analogously, $T_k$ represents the probability of obtaining a random pair of patterns belonging to the same cluster in $C$ and $C'$ simultaneously. A higher value of $B_k$ indicates higher consistency between both clustering solutions. A value of $B_k = 1$ indicates that $C$ and $C'$ are identical clustering solutions, whereas a value of $B_k = 0$ indicates that no pair of elements can be found belonging to the same cluster in $C$ and $C'$ simultaneously.


\vspace{-4mm}
\section{Results and discussion}\label{sec:results}
The proposal has been applied to both real biological datasets, YEAST and ARA. For the experiments, in each dataset, a random subsampling of 90\% of the total genes was performed at the beginning in order to obtain set $A$ (genes with well-known biological function); for the remaining 10\% of the genes, semantic-based information was artificially removed in order to use this set as $B$ (genes with unknown biological function). The information removed from $B$ is used at the end of the approach for validation of results.

The source code is freely available for academic use at http://sourceforge.net/projects/sourcesinc/files/gamma\\AM/1.0/. 
A user-friendly access is provided as web interface at http://fich.unl.edu.ar/sinc/web-demo/gamma-am/. 

\subsection{YEAST results}
For YEAST data, both expression distance and semantic distance matrices were built. Their corresponding histograms were calculated (see Figure \ref{fig:02}) and since there is a high imbalance between them, both available methods (\disc{} and $\gamma$-tuning method) to obtain more uniform matrix distributions were applied. For the \disc{} method, a value of $m = 20$ intervals was used. The resulting histograms are shown in Figure \ref{fig:03} after application of the \disc{} method for the Euclidean and Relevance distances\footnote{Pearson correlation, Lin and Resnik measures show similar results.}. Usually, genes with unknown biological function are present in different proportions within a given genome. Therefore, several proportions for $A_1/A_2$ were used to test the robustness of the method in cases when there are different proportions of genome annotations. All tested proportions yielded similar results, thus the proportion $A_1/A_2 = 50/50$ was selected. In order to select an appropriate $\gamma$ value, SC was calculated and averaged through 10 runs for each value of $\gamma$ and the results have been shown in Figure 4. 
It can be seen that the best compactness value is 0.766 ($\gamma = 0.95$). Therefore, that value of $\gamma$ will be used as a parameter to measure the contributions of the expression and the semantic GO-based information in the gamma distance

In Step 1, the resulting pairwise distances provided by the two methods presented above (\disc{} and $\gamma$-tuning) were used to obtain the partition $G^{A_2 \rightarrow A_1}$.
For clustering, since a very high value of $k$ would assign each object to a single cluster, whereas a very low value would cause the clusters to be excessively large, a value of $k=10$ was used according to the Gap statistic\footnote{Gap estimates the adequate number of clusters for a dataset.} \cite{tibshirani2001estimating} and a value of 10 was selected for the number of runs of the clustering algorithm.

Finally, the partition $G^{A_2 \rightarrow A_1}$ was measured in terms of biological quality upon $BHI$ and $BC$. Results for the $\gamma$-tuning (using the best $\gamma$ value found according to the proportion $50/50$ and the $SC$ measure, see Figure 6) and \disc{} method are shown in Table \ref{tab:results-bhi-gamma-tuning-yea}. The first column indicates the method used. For the percentile method, results of its application are shown in the second row (Yes) and results obtained without its application in the third row (No). In these cases, for this method, a value of $\gamma = 0.50$ has been used. From the analysis of Table \ref{tab:results-bhi-gamma-tuning-yea}, it can be seen that both $BHI$ and $BC$ measures indicate $\gamma$-tuning (with its corresponding appropriate $\gamma$ value) as a better method than the \disc{} one for obtaining a better final partition, in terms of better biological quality of the results. It should be noted that although $\gamma$-tuning and no application of the \disc{} method have the same $BHI$ value, the biological compactness is better for the final partition when the $\gamma$-tuning method has been used to determine the appropriate $\gamma$ to be used to balance the expression and semantic GO-based data. 

\begin{table}[t!]
\centering
\begin{tabular}{c|c|c|c}
\multicolumn{ 2}{c|}{\multirow{2}{*}{Method}} & 
\multicolumn{ 2}{c}{$G^{A_2 \rightarrow A_1}$} \\ 
\cline{ 3- 4} 
\multicolumn{1}{l}{} & \multicolumn{1}{l|}{} & \multicolumn{ 1}{c|}{BHI} & \multicolumn{1}{c}{BC} \\ \hline
\multicolumn{ 2}{c|}{\multirow{1}{*}{$\gamma$-tuning}} & 0.21 & 0.50 \\ 
\hline \hline 
\multicolumn{ 1}{c|}{\multirow{2}{*}{\Disc{}}} & \multicolumn{ 1}{c|}{Yes} 	& 0.14	& 0.60	\\
 & \multicolumn{ 1}{c|}{No}   & 0.21	& 0.52	\\
\end{tabular}
\vspace{\tfill{}}
\caption{YEAST dataset. BHI and BC after applying the $\gamma$-tuning method and the \disc{} method.}
\label{tab:results-bhi-gamma-tuning-yea}
\vspace{\tffill{}}
\end{table}

Thus, for the Step 2 of our approach, in order to obtain the final partition $G^{B \rightarrow A}$, the value of $\gamma= 0.95$ was used. The biological quality of the resulting partition has been measured with $BHI$ and $BC$ and results are reported in Table \ref{tab:results-test-yea}. 
In this table, the first row shows the results of \dif{a state-of-the-art approach for this problem, proposed by Rhee \emph{et al.}}  (as described in \cite{rhee2008use}) where the complete dataset $A \cup B$ is considered for expression-based only clustering ($\gamma = 0$). 
The second row shows the performance of clustering the input dataset $A$ with $\gamma = 1$, and assigning B genes with $\gamma = 1$, thus considering full semantic information (as the best possible case). 
The third row shows the output for the \gss{} method as described in Section \ref{sec:02}.  
An ANOVA test was performed on these results over 100 runs for each dataset ($\alpha = 0.05$) showing significant difference among methods. 
It can be seen that the results obtained by the \gss{} approach using a suitable $\gamma$ value automatically obtained by $\gamma$-tuning for balancing the original expression and semantic GO-based data, has high biological quality according to both $BHI$ and $BC$ indexes. 
In the case of $BHI$, $G^{B \rightarrow A}$ is equal to the 0.19 value for this index in this dataset if full biological (GO-based) semantic data were considered for clustering genes. In the case of $BC$, a similar conclusion can be drawn: while the application of clustering with complete information yields the best possible biological compactness of 0.58, the $G^{B \rightarrow A}$ partition presents an equivalent value, providing significantly better results when compared to the classical approach \cite{rhee2008use} where no semantic information is used at all for clustering original expression data. \diff{This result is a major improvement in obtaining biologically meaningful clusters, since the application of the gamma distance without having complete semantic-based information yields similar results to the measures obtained using semantic-based data only. Therefore, through the application of our proposal, clusters formed with incomplete information can still show the same biological quality than those obtained with full information.}

In order to evaluate the quality of our approach but only \dif{regarding the clustering results (Step 2), the \gss{} approach has been compared with the related work \cite{brameier2007co}\footnote{This approach does not infer GO terms, thus it can be partially compared just to the clustering part of Step 2}, where a SOM is used to co-cluster gene expression and semantic data. In this experiment, genes from the complete dataset  $A \cup B$  were clustered considering expression-based and semantic-based data using the optimum gamma value determined previously for this dataset. Results for this comparison are shown in Table \ref{tab:results-comparison-yea}. The second and third row show results for the SOM-based approach in two  configurations. The first row shows results for the \gss{} approach. $BHI$ and $BC$ measures have been calculated for each case.}
\dif{It can be seen that our clustering procedure outperforms the SOM-based method in all cases with the best biological homogeneity (0.19) against the SOM-based method (0.06 and 0.04). Analogously with $BC$, the compactness in terms of biological data is the best for our method (0.58) when compared to the SOM-based method (0.78 and 0.80)}
\dif{These results show that the clustering procedure performed in Step 2 is capable of obtaining high quality partitions when compared to other state-of-the-art approach, with the same input data.}
\diff{Moreover, the above 100\% increase in biological quality is a strong evidence of the strength and importance of this step in the \gss{} approach.}

In order to evaluate the quality of the Step 2 or our approach with respect to the assignment of B genes to clusters, the FM index has been used. This index evaluates the similarity between two clustering solutions, where higher values indicate high similarity between clustering solutions. The comparison was performed between what would be a random assignment of genes to clusters, and our method, versus the best possible case ($\gamma = 1$), calculating the corresponding FM index for each scenario. A value of 0.12 was obtained when comparing the random assignment against the best possible case, whereas a higher value of 0.19 was obtained when comparing our method to the best possible case. This result represents a percentual improvement of above 60\% regarding closeness between the partition obtained by our assignment procedure and the partition that would be obtained with full semantic information.

For Step 3 of our approach, once the partition was obtained with the selected $\gamma$, an enrichment analysis procedure was applied using the \emph{g:Profiler} tool \cite{reimand2011g}. Each cluster was characterized with biological functions\footnote{Full results are provided in Supplementary Material 2.}.
Only those genes from each cluster which originally belonged to A were selected as input to the enrichment. The full YEAST dataset was used as background gene set. 
An example is described in Table \ref{tab:enrichment-analysis}, where the results of the enrichment of cluster \clusteryea{} are shown. The $p$-value for each GO term corresponding to the input genes was calculated, and only those terms with a $p$-value lower than a threshold (0.05 commonly used in literature) were considered meaningful \cite{reimand2011g,subramanian2005gene}. These terms were ordered according to their $p$-value and their corresponding functions were assigned as the biological functions of the unknown B genes present in cluster \clusteryea{}. 

\dif{To evaluate the quality and effectiveness in inferring genomic annotation of Step 3 of the proposed method, the recall of the labels inferred for the B genes is reported in Table \ref{tab:accuracy_yea}. This recall was calculated as the matching proportion of inferred GO terms with respect to original terms, averaged for all B genes.} It has to be noticed here that, for calculating this measure, the most popular GO terms used to annotate most genes in each dataset can introduce a bias and unusually high values can be obtained as results of the characterization, even when inferring gene function just by chance. Such insight for the YEAST dataset is provided in Table \ref{tab:count-etiquetas}. The first and second columns describe the GO label. The third column shows the number of genes in the dataset annotated with the corresponding GO label. It becomes clear from the table that only two of the GO labels (translation and cytoplasmic translation) annotate most of the genes in the entire dataset (we named those labels as popular GO labels). It can be clearly noticed here the high imbalance in the distribution of GO label annotations among the top annotating labels in the dataset and the rest of the GO labels. Thus, there is a high probability of finding those as matching labels between the inferred and the original biological functions for the B genes.


\begin{table}[t!]
\centering
\begin{tabular}{l|c|c}
\multicolumn{ 1}{c|}{Method} & BHI & BC \\ 
\hline 
Classical \cite{rhee2008use} & 
\multicolumn{ 1}{c|}{\multirow{1}{*}{0.07}} & 
\multicolumn{ 1}{c}{\multirow{1}{*}{0.76}} \\ 
\hline
Complete information & 0.19 & 0.58 \\ 
\hline
\gss{} method & 0.19 & 0.58 \\ 
\end{tabular} 
\vspace{\tfill{}}
\caption{Quality results for \gss{} on the YEAST dataset.}
\label{tab:results-test-yea}
\vspace{\tftfill{}}
\end{table}

\begin{table}[t!]
\centering
\begin{tabular}{l|c|c}
\multicolumn{ 1}{c|}{Method} & BHI & BC \\ 
\hline
SOM \cite{brameier2007co}, $m=0$ & 0.06 & 0.78 \\ 
\hline
SOM \cite{brameier2007co}, $m=1$ & 0.04 & 0.80 \\ 
\hline 
\gss{} method & 0.19 & 0.58 \\ 
\end{tabular} 
\vspace{\tfill{}}
\caption{Comparison on clustering quality only for Step 2, YEAST dataset.}
\label{tab:results-comparison-yea}
\vspace{\tftfill{}}
\end{table}

\begin{table}[t!]
\centering
\scriptsize
\setlength\tabcolsep{2pt} 
\begin{tabular}{r|c|l}
\textbf{$p$-value} & \textbf{GO term} & \textbf{Description} \\ \hline
 3.46e-57 & GO:0032774 & RNA biosynthetic process \\
 3.46e-57 & GO:0097659 & nucleic acid-templated transcription \\
 3.46e-57 & GO:0006351 & transcription, DNA-templated \\
 1.67e-47 & GO:0034654 & nucleobase-containing compound biosynthetic process \\
 5.16e-45 & GO:0019438 & aromatic compound biosynthetic process \\
 2e-44 & GO:0018130 & heterocycle biosynthetic process \\
 2.77e-43 & GO:0044271 & cellular nitrogen compound biosynthetic process \\
 3.03e-43 & GO:2001141 & regulation of RNA biosynthetic process \\
 3.03e-43 & GO:1903506 & regulation of nucleic acid-templated transcription \\
 3.03e-43 & GO:0006355 & regulation of transcription, DNA-templated \\
 \end{tabular}
\vspace{0.5mm}
\caption{Output of enrichment analysis for cluster \clusteryea{} for YEAST dataset.}
\label{tab:enrichment-analysis}
\vspace{-4mm}
\end{table}

\begin{table}[t!]
\centering
\scriptsize
\begin{tabular}{c|c|c}
GO name & GO ID & Annotation count \\ 
\hline 
translation & GO:0006412 & 216 \\ 
\hline 
cytoplasmic translation & GO:0002181 & 201 \\ 
\hline 
rRNA processing & GO:0006364 & 36 \\ 
\hline 
ribosome biogenesis & GO:0042254 & 36 \\ 
\hline 
(other GO labels) & (\ldots) & $<$ 30 \\ 
\end{tabular} 
\vspace{\tfill{}}
\caption{GO label count on the YEAST dataset.}
\label{tab:count-etiquetas}
\vspace{\tftfill{}}
\end{table}

Therefore, the recall has been calculated with and without the popular GO labels. The results are shown in Table \ref{tab:accuracy_yea}. The first row shows performance for the \classi{} classical approach, the second row shows performance considering full semantic information, the third row shows the performance obtained when applying our method. It can be seen in the second column that the recall of 0.218, obtained with our method on the original (biased) distribution of GO labels, is considerably higher than the low value (0.019) obtained by the classical approach, and relatively close to the best possible value (0.248). The relationship among these values was calculated as percentual difference with respect to the complete information method, and shown in the third and fifth column. Results obtained with our method differ in only 12.10\% to the best possible case. Instead, the classical approach has a percentual error of 92.34\% with respect to the best case. When cutting off popular GO labels, these relations between those methods remain in a fairly similar way, as shown in the fourth column. It can be seen that the recall difference of the inferred labels with respect to the best case is very similar in comparison to the classical approach, also when the measure is calculated without popular terms. These results indicate the robustness of our method to the problem of an imbalanced distribution of GO annotations. \diff{In addition to this, the remarkable improvement of 80\% when compared to the classical approach for label inference proves the effectiveness of our proposal.}

\begin{table}[t!]
\setlength{\tabcolsep}{2pt} 
\renewcommand{\arraystretch}{1} 
\centering 
\scriptsize
\begin{tabular}{l|c|c|c|c}
\multicolumn{ 1}{c|}{\multirow{2}{*}{Method}} & \multicolumn{ 1}{c|}{\multirow{2}{*}{Recall}} &  Relative & Recall without & Relative \\ 
 &  & Error     & popular GO labels     & Error \\
\hline
Classical \cite{rhee2008use} & \multicolumn{ 1}{c|}{\multirow{1}{*}{0.019}} & \multicolumn{ 1}{c|}{\multirow{1}{*}{-92.34\%}} &
\multicolumn{ 1}{c|}{\multirow{1}{*}{0.003}} & \multicolumn{ 1}{|c}{\multirow{1}{*}{-93.88\%}}\\ 
\hline 
Complete information & \multicolumn{ 1}{c|}{\multirow{1}{*}{0.248 }} & \multicolumn{ 1}{c|}{\multirow{1}{*}{- }} & \multicolumn{ 1}{c|}{\multirow{1}{*}{0.049}} & \multicolumn{ 1}{c}{\multirow{1}{*}{-}} \\
\hline
\gss{} method & 0.218 & -12.10\% & 0.038 & -22.45\% \\ 
\end{tabular} 
\vspace{\tfill{}}
\caption{Recall for the YEAST dataset. Error is the percentual difference with respect to complete information for each method.}
\label{tab:accuracy_yea}
\vspace{\tffill{}}
\end{table}

\begin{sidewaysfigure*}
\centerline{\includegraphics[width=0.8\textwidth]{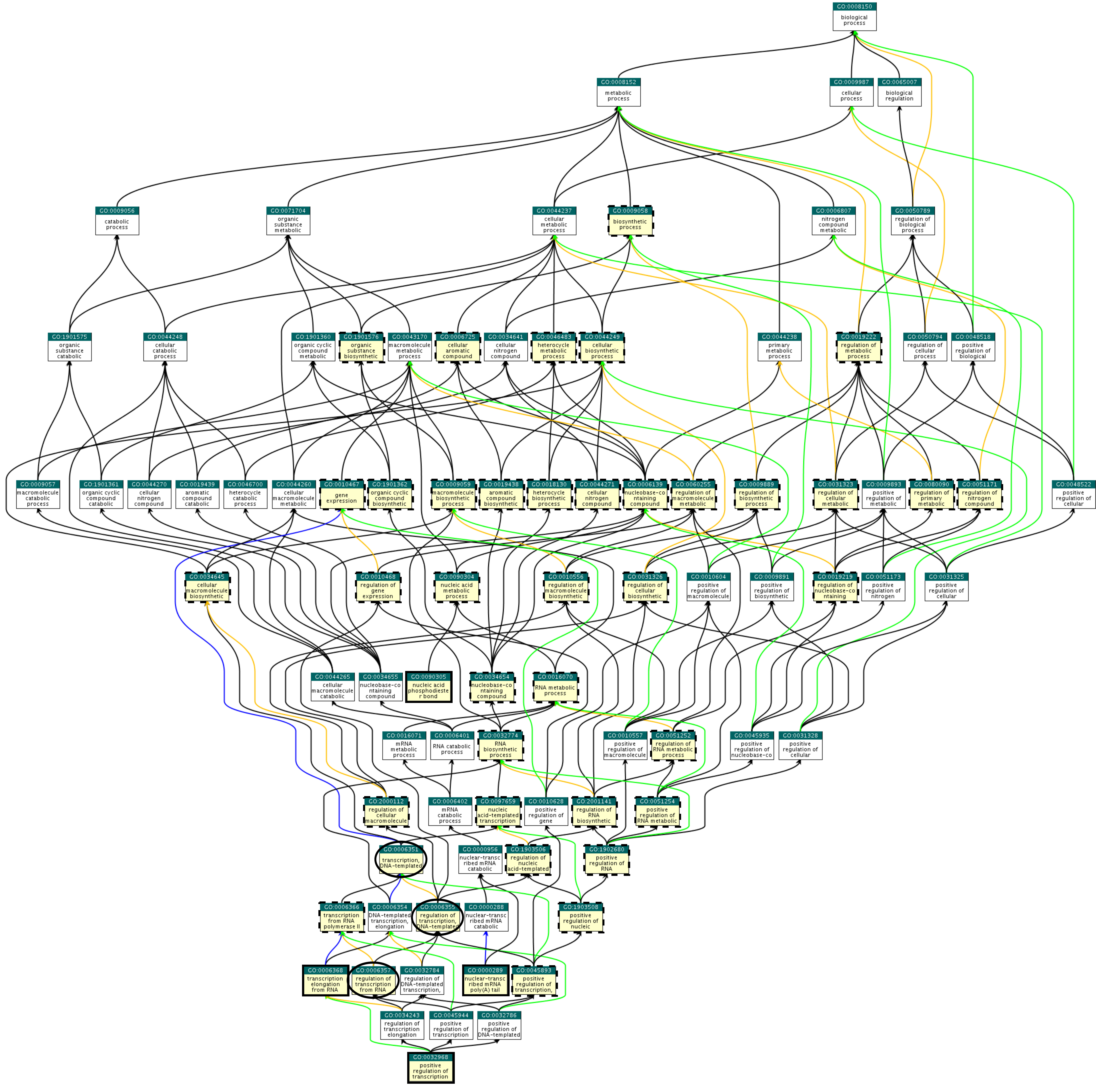}}
\caption{Inferred terms in the Gene Ontology for gene \geneexampleyea{} from cluster \clusteryea{} of the YEAST dataset: assigned terms (dashed line), matching terms (ellipse), original B terms (thick line).} \label{fig:mapa_go}
\end{sidewaysfigure*}

In order to further validate the inferred biological functions for the $B$ genes, comparison was performed with the already well-known semantic information of the $B$ genes, which was removed at the beginning of the experiments. An example of this is shown in Table \ref{tab:validation-assignment-procedure}. Some of the genes from $B$ which were assigned to cluster \clusteryea{} are shown in the first column. Actual GO terms corresponding to those genes are shown in the second column. Terms that match exactly the ones assigned by the proposed method in this work are shown in bold. It can be clearly seen from this example that one or more than one GO terms have been perfectly inferred by the proposed procedure. In fact, this can be seen graphically in Figure \ref{fig:mapa_go}. Relative location of inferred terms in GO is shown in the figure. Terms assigned to B that match the actual terms are highlighted by an ellipse. The remaining terms resulting from the enrichment analysis procedure are shown with a dashed line. Original terms from B are surrounded by a dark thick line. It can be seen that there is a closeness in terms of semantic relationship among these terms. Furthermore, matching terms from B are always directly related to assigned terms, both as parents and siblings. In particular, many of the children from matching terms are actually inferred terms. Besides, original terms from B are very close to assigned terms. In summary, inferred terms match perfectly, or are related as parents or siblings of actual $B$ terms. 
\vspace{-4mm}

\begin{table}[t!]
\centering 
\scriptsize
\begin{tabular}{l|lll}
\multicolumn{ 1}{c|}{\textbf{Gene}} & \multicolumn{2}{c}{\textbf{GO actual terms}} \\ \hline
 POP2  
 & \textbf{GO:0006351}  & GO:0032968 \\
 & \textbf{GO:0006355}  & GO:0090305 \\
 & \textbf{GO:0006357} & GO:0000289 \\
 & GO:0006368 & \\

NTG1   
& \textbf{GO:0008152} & GO:0000737 \\
& GO:0006281  & GO:0006284 \\
 & GO:0034599 & GO:0090297 \\
 & GO:0006974  & GO:0006285 \\

%
%

\end{tabular}
\vspace{\tfill{}}
\caption{Validation results for some genes from B assigned to cluster \clusteryea{} for YEAST dataset. Original matching labels shown in bold}
\label{tab:validation-assignment-procedure}
\vspace{\tffill{}}
\end{table}

\subsection{ARA results}
Regarding ARA dataset, an analogous procedure to the YEAST dataset was performed. 
As shown in Tables \ref{tab:results-test-ara}, \ref{tab:results-comparison-ara} and \ref{tab:accuracy_ara}, similar results to the YEAST dataset are achieved, thus equivalent conclusions can be drawn. Due to space restrictions, full results and analysis can be found in Supplementary Material 1.

\dif{Clustering performance was also evaluated for this dataset, comparing the clustering in Step 2 to the SOM-based co-clustering in terms of $BHI$ and $BC$. Results are shown in Table \ref{tab:results-comparison-ara}. It can be seen that the clustering quality of \gss{} outperforms the SOM co-clustering for this dataset as well. Our method yields the best $BHI$ (0.12) compared to the lower values (0.06 and 0.10) for the SOM-based method. Similar results are reached in terms of $BC$, with the best value (0.59) against the less compact results for the SOM approach.}

Regarding biological quality, Table \ref{tab:results-test-ara} shows the results of our method in the third row. The second row shows the results for the best possible case, using complete semantic information, whereas the first row shows the values for the \classi{} classical approach. It can be seen that, for both measures, the application of the $\gamma$-tuning method with an appropriate $\gamma$ value used to balance the amount of the expression and semantic GO-based data is significantly close in terms of biological quality to the best possible values of both performance measures. For $BHI$, the value of 0.14 obtained with our method is remarkably near to the best value of 0.15, representing a major improvement to the value of 0.05 obtained when applying the classical method. Similar results are obtained for BC, where a value of 0.60 for our method is very close to the best value of 0.59 obtained with full semantic information and indicates a considerable higher compactness than the value of 0.82, which was obtained with $A \cup B$ with $\gamma = 0$. 
Recall was also calcultated for this dataset, as shown in Table \ref{tab:accuracy_ara}. It can be seen that the value of 0.052 obtained with \gss{} is very close to the best possible value of 0.071, and fairly better than the value of 0.033 obtained by the classic approach. These analysis stands also when popular terms are not taken into account.
The quality of Step 2 was also evaluated with the FM index. A random assignment and our approach were compared against the best possible case, obtaining the values 0.12 and 0.15, respectively. These results show a similarity improvement of nearly 25\%  between the partition obtained with \gss{} and the partition that would be obtained with complete semantic information.

\begin{table}[t!]
\centering
\begin{tabular}{l|c|c}
\multicolumn{ 1}{c|}{Method} & BHI & BC \\ 
\hline 
Classical \cite{rhee2008use} & 
\multicolumn{ 1}{c|}{\multirow{1}{*}{0.05}} & 
\multicolumn{ 1}{c}{\multirow{1}{*}{0.82}} \\ 
\hline
Complete information & 0.15 & 0.59 \\ 
\hline
\gss{} method & 0.14 & 0.60 \\ 
\end{tabular} 
\vspace{\tfill{}}
\caption{Quality results for \gss{} on the ARA dataset.}
\label{tab:results-test-ara}
\vspace{\tftfill{}}
\end{table}

\begin{table}[t!]
\centering
\begin{tabular}{l|c|c}
\multicolumn{ 1}{c|}{Method} & BHI & BC \\ 
\hline
SOM \cite{brameier2007co}, $m=0$ & 0.06 & 0.74 \\ 
\hline
SOM \cite{brameier2007co}, $m=1$ & 0.10 & 0.82 \\ 
\hline 
\gss{} method & 0.12 & 0.59 \\ 
\end{tabular} 
\vspace{\tfill{}}
\caption{Comparison on clustering quality only for Step 2, ARA dataset.}
\label{tab:results-comparison-ara}
\vspace{\tftfill{}}
\end{table}

\setlength{\tabcolsep}{2pt} 
\renewcommand{\arraystretch}{1} 
\begin{table}[t!]
\centering 
\scriptsize
\begin{tabular}{l|c|c|c|c}
\multicolumn{ 1}{c|}{\multirow{2}{*}{Method}} & \multicolumn{ 1}{c|}{\multirow{2}{*}{Recall}} &  Relative & Recall without & Relative \\ 
 &  & Error     & popular GO labels     & Error \\
\hline
Classical \cite{rhee2008use} & 
\multicolumn{ 1}{c|}{\multirow{1}{*}{0.033}} & \multicolumn{ 1}{c|}{\multirow{1}{*}{-53.52\%}} &
\multicolumn{ 1}{c|}{\multirow{1}{*}{0.034}} & \multicolumn{ 1}{|c}{\multirow{1}{*}{-27.66\%}}\\ 
\hline 
Complete information &
\multicolumn{ 1}{c|}{\multirow{1}{*}{0.071}} & \multicolumn{ 1}{c|}{\multirow{1}{*}{-}} &
\multicolumn{ 1}{c|}{\multirow{1}{*}{0.047}} & \multicolumn{ 1}{|c}{\multirow{1}{*}{-}}\\ 
\hline
\gss{} method & 0.052 & -26.76\% & 0.038 & -19.15\% \\ 
\end{tabular} 
\vspace{\tfill{}}
\caption{Recall for the ARA dataset.}
\label{tab:accuracy_ara}
\vspace{\tffill{}}
\end{table}

In summary, as it can be seen from the results in both real datasets from different species, it can be stated that our proposal has effectively succeeded in assigning biological function to non-annotated genes. This is supported by the results obtained from both the performance and validation measures considered, as well as the graphical analysis of the annotations inferred for the B genes.
\diff{Therefore, our proposal establishes an important milestone regarding computational methods to take full advantage of an external source of biological information, with the aim of discovering gene functions.}

\vspace{-4mm}
\section{Conclusions}\label{sec:conclusions}
A novel approach for inferring biological function for a set of unknown genes has been presented in this work. It is based on the assignment of unknown genes to groups of genes with well-known information, and the application of enrichment analysis to the groups in order to characterize those unknown genes. It has been tested on two real datasets from different species, and compared to the state-of-the-art clustering approach, obtaining very good results that prove the effectiveness of the proposal. These results show the convenience of using \gss{} to infer biological knowledge from a set of genes with unknown biological function. The proposed approach can constitute an important starting point for guiding biologists into the inference of possible function to recently discovered genes, as well as to the design of the most adequate wet experiments to further confirm their functional behaviour. The approach can lead the biologist into a convenient path through the large GO-BP structure, which would help finding the correct biological function to genes with previously unknown semantic information.

\vspace{-4mm}
\section*{Acknowledgements}
This work was supported by National Scientific and Technical Research Council (CONICET) [PIP 2013-2015 117] and Agencia Nacional de Promoci\'{o}n Cient\'{i}fica y Tecnol\'{o}gica (ANPCyT) [PICT 2014 2627].

\bibliographystyle{IEEEtran}
\vspace{-4mm}
\bibliography{IEEEabrv,bibliography}

\vspace*{\bfill{}}

\begin{IEEEbiographynophoto}{G. Leale}
is Teaching Assistant in UTN-FRRo. He is currently a Ph.D. student at the National Scientific and Technical Research Council (CONICET), Argentina. His research interests are in the fields of computational intelligence, data mining and bioinformatics.
\end{IEEEbiographynophoto}

\vspace*{\bfill{}}

\begin{IEEEbiographynophoto}{A. Baya}
is member of the CIFASIS  Research Institute, part of CONICET. He holds the position of Assistant Researcher. His main research  interests are clustering  methods, clustering validation methods, and their applications to biological data.
\end{IEEEbiographynophoto}

\vspace*{\bfill{}}

\begin{IEEEbiographynophoto}{P. Granitto}
is a full-time Researcher at CONICET and UNR. He leads the Machine Learning Group at CIFASIS. His current  research interests include application of modern machine learning techniques to agroindustrial and biological problems.
\end{IEEEbiographynophoto}

\vspace*{\bfill{}}

\begin{IEEEbiographynophoto}{D.H. Milone}
is Full Professor in National University of Litoral (UNL) and Research Scientist at CONICET. His research interests include statistical learning, pattern recognition, signal processing, with applications to biomedical signals and bioinformatics.
\end{IEEEbiographynophoto}

\vspace*{\bfill{}}

\begin{IEEEbiographynophoto}{G. Stegmayer}
is Assistant Professor in UNL and full-time Adjunct Researcher at CONICET. Her current research interests involve machine learning, data mining and pattern recognition in bioinformatics.
\end{IEEEbiographynophoto}

\clearpage

\section*{Inferring unknown biological functions by integration of GO annotations and gene expression data. \\ Supplementary Material}

\subsection*{ARA results}
\begin{figure}[ht]
\centerline{\includegraphics[scale=0.23]{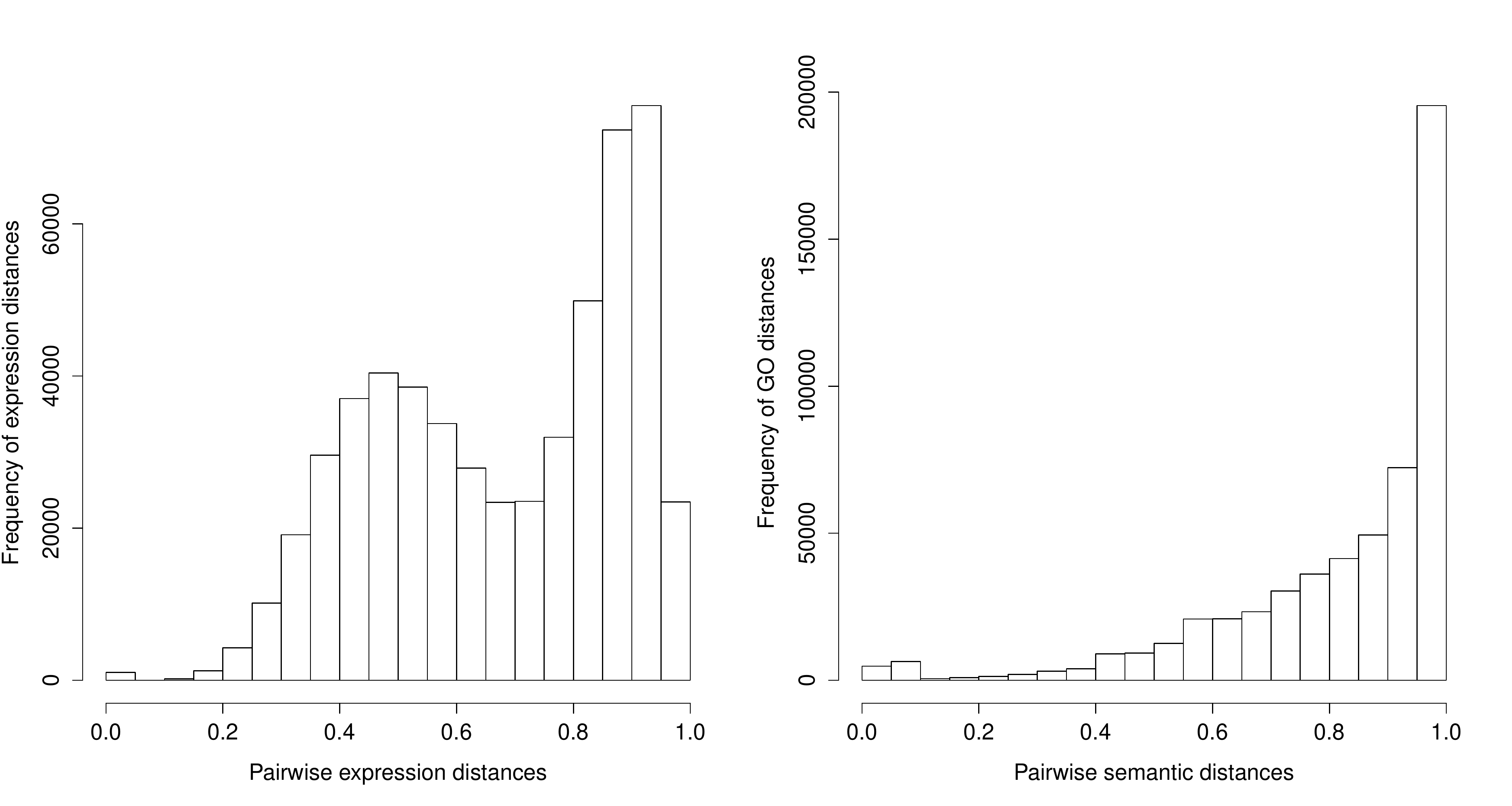}}
\caption{Original distance histograms distributions for the ARA gene dataset. Expression-based pairwise distances (left) and semantic-based pairwise distances (right).}\label{fig:02-ara}
\end{figure}

\begin{figure}[t]
\centerline{\includegraphics[scale=0.23]{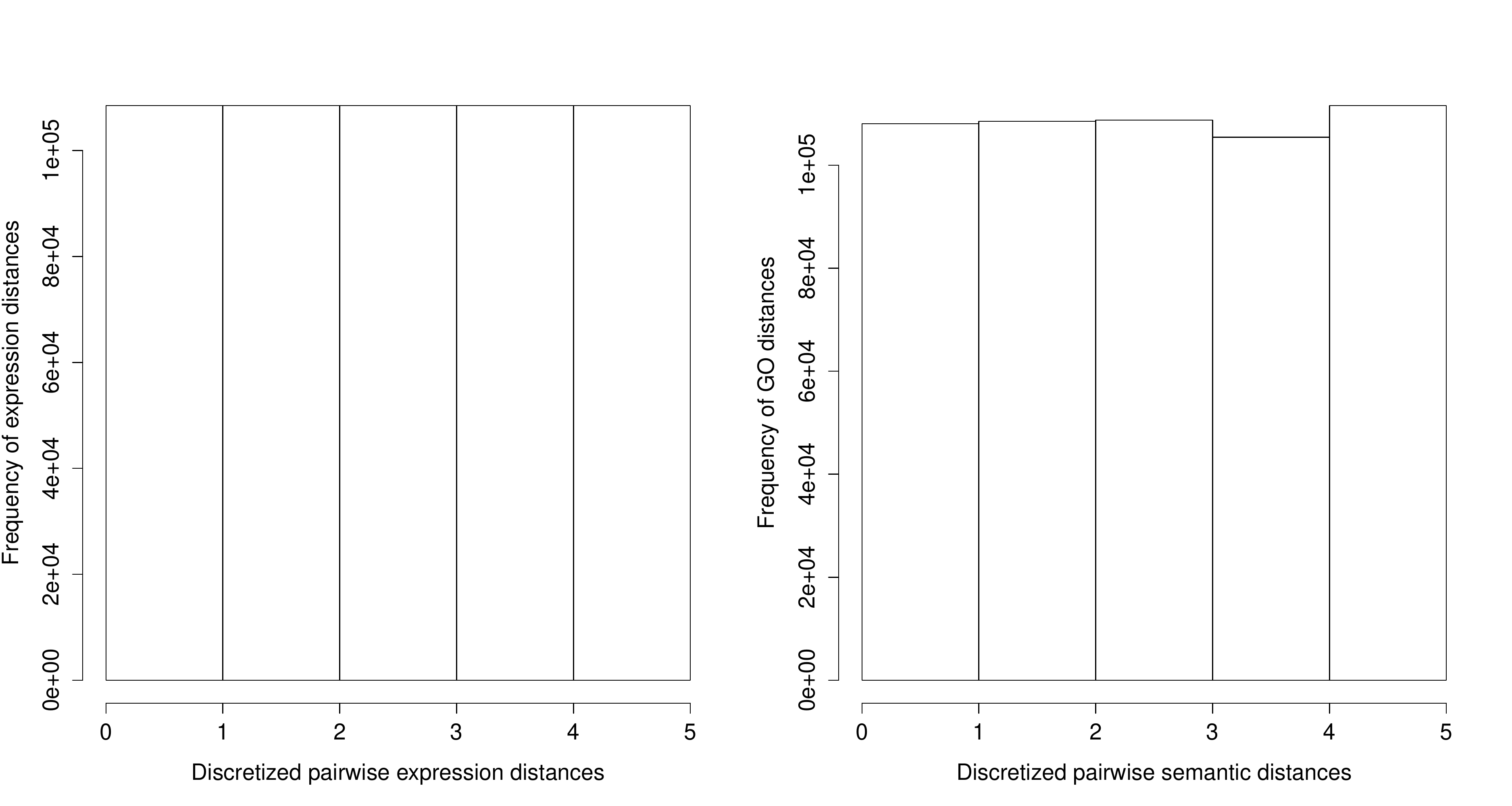}}
\caption{Histograms for the ARA pairwise distances from Figure \ref{fig:02-ara} after applying the \disc{} method. Expression-based pairwise distances (left) and semantic-based pairwise distances (right).}\label{fig:03-ara}
\end{figure}

\begin{figure*}[t!]
\centerline{\includegraphics[scale=0.48,trim=0 20 0 50, clip]{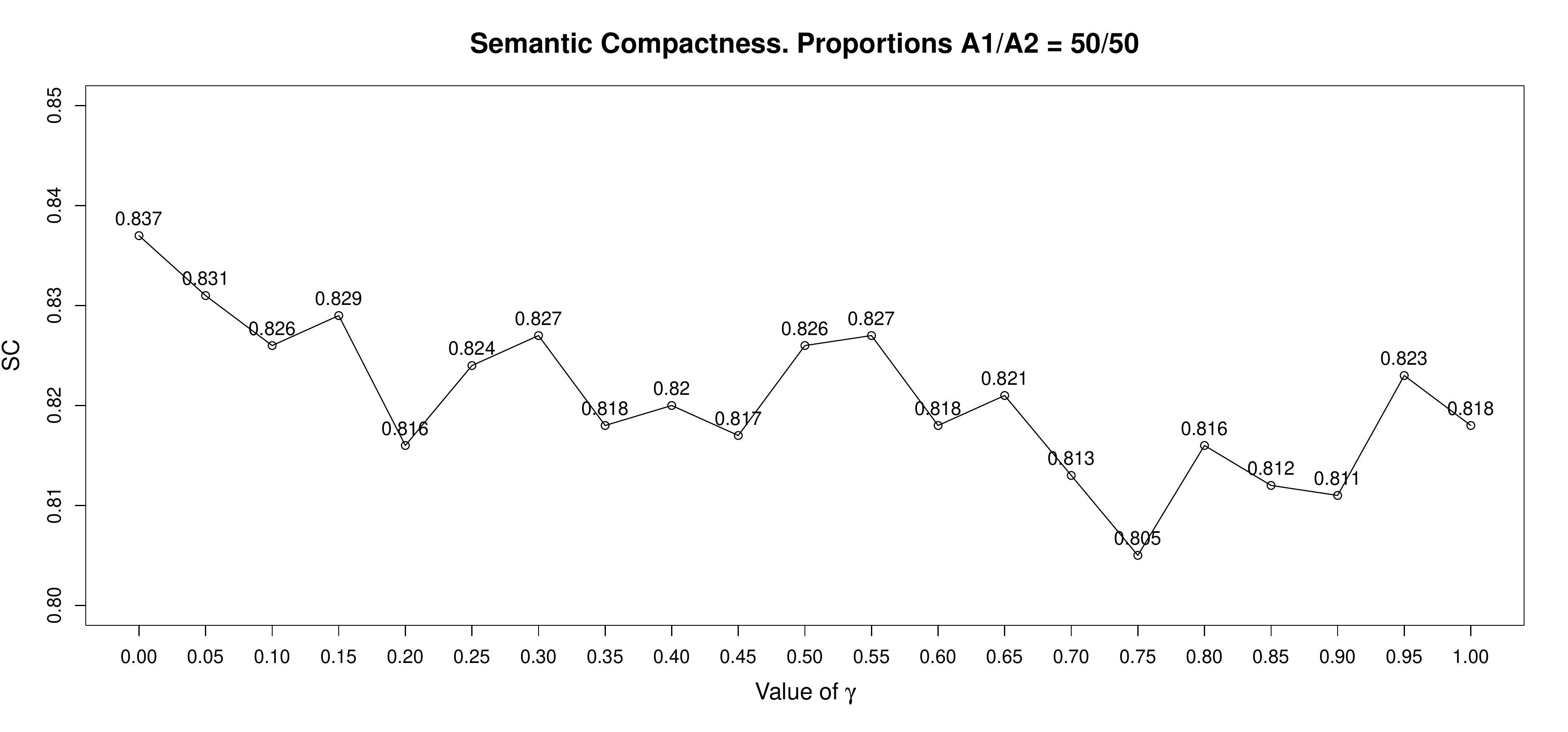}}
\caption{Results for the $\gamma$-tuning method on the ARA dataset.} \label{fig:results-gamma-measure-ara}
\end{figure*}

For ARA dataset, an analogous procedure to the YEAST dataset was performed. Expression distance and semantic distance matrices were built and both the \disc{} and the $\gamma$-tuning method were applied. Histograms for expression distance and semantic distance matrices are shown in Figure \ref{fig:02-ara} before applying the \disc{} method and in Figure \ref{fig:03-ara} after the application of this method. Application of the $\gamma$-tuning method is shown in Figure \ref{fig:results-gamma-measure-ara}.

Measures for validation were also calculated on this dataset. Results for $BHI$ and $BC$ on the partition $G^{A_2 \rightarrow A_1}$ are shown in Table \ref{tab:results-bhi-gamma-tuning-ara}. The first column indicates the method applied. For the $\gamma$-tuning method, which is shown in the first row, an appropriate $\gamma$ value of 0.75 was used for a proportion of $A1/A2 = 50/50$, as resulting from Figure \ref{fig:results-gamma-measure-ara}. Results obtained when applying the \disc{} method are shown in the second row (Yes), and when not applying this method are shown in the third row (No). In both cases a gamma value of $\gamma = 0.50$ was used.

It can be seen from the table that, for both measures, the $\gamma$-tuning method clearly provides the best results. This shows that higher biological quality is reached on the partition $G^{A_2 \rightarrow A_1}$ with this method when it is compared to the \disc{} method, analogously to the performance measures obtained on the YEAST dataset.

In order to evaluate the clustering quality of Step 2, recall was also calculated for this dataset. An annotation count has also been calculated in this case, as shown in Table \ref{tab:count-etiquetas-ara}. The first and second columns describe the GO label, and the third column shows the number of genes in the dataset annotated with the corresponding GO label. As it can be seen in the table, there is not a noticeable difference among the label counts, and therefore a more homogeneus label distribution is found in this case, with no distinguishable popular labels to consider. It should be noted here that this is probably due to the fact that Arabidopsis is a model specie and it is thoroughly studied in literature, and therefore its GO terms are well annotated. Thus, the recall calculation should not be very much affected by the removal of popular terms. Anyway, the first three most used labels (regulation of transcription, DNA-templated, response to salt stress and oxidation-reduction process) were removed to make a comparison against the full label set. Results are shown in Table \ref{tab:accuracy_ara}. The first row shows recall for the classic approach, while the second row shows recall considering complete semantic information, and the third row shows the recall with our approach. In the second column, it can be seen that the value of 0.052 obtained with \gss{} is very close to the best possible value of 0.071, and fairly better than the value of 0.033 obtained by the classic approach. Columns comparing these values by calculating differences with respect to the complete information case are also provided in the third column. Here, while our method has a percentual error of 26.76\% with respect to the best case, the classical approach has the double of percentual errors in assigning labels to unknown genes. Recall without taking into account the first labels (fourth column and fifth columns) show similar results, proving the effectiveness of our method when applied to this dataset as well. 

\begin{table}[t!]
\centering
\begin{tabular}{c|c|c|c}
\multicolumn{ 2}{c|}{\multirow{2}{*}{Method}} & 
\multicolumn{ 2}{c}{$G^{A_2 \rightarrow A_1}$} \\ 
\cline{ 3- 4} 
\multicolumn{1}{l}{} & \multicolumn{1}{l|}{} & \multicolumn{ 1}{c|}{BHI} & \multicolumn{1}{c}{BC} \\ \hline
\multicolumn{ 2}{c|}{\multirow{1}{*}{$\gamma$-tuning}} & 0.20 & 0.50 \\ 
\hline \hline 
\multicolumn{ 1}{c|}{\multirow{2}{*}{\Disc{}}} & \multicolumn{ 1}{c|}{Yes} 	& 0.14	& 0.56\\
 & \multicolumn{ 1}{c|}{No}   & 0.08	& 0.67		\\
\end{tabular}
\vspace{\tfill{}}
\caption{ARA dataset. BHI and BC after applying the $\gamma$-tuning method and the \disc{} method.}
\label{tab:results-bhi-gamma-tuning-ara}
\end{table}

\begin{table}[t!]
\centering
\scriptsize
\begin{tabular}{c|c|c}
GO name & GO ID & Annotation count \\ 
\hline 
regulation of transcription,& 
\multicolumn{ 1}{c|}{\multirow{2}{*}{GO:0006355}} & 
\multicolumn{ 1}{c}{\multirow{2}{*}{137}} \\ 
DNA-templated & & \\
\hline 
response to salt stress & GO:0009651 & 97 \\ 
\hline 
oxidation-reduction process & GO:0055114 & 95 \\ 
\hline 
metabolic process & GO:0008152 & 87 \\ 
\hline 
response to abscisic acid & GO:0009737 & 70 \\ 
\hline 
response to cadmium ion & GO:0046686 & 66 \\ 
\hline 
response to cold & GO:0009409 & 60 \\ 
\hline 
(other GO labels) & (\ldots) & $<$ 60 \\ 
\end{tabular} 
\vspace{\tfill{}}
\caption{GO label count on the ARA dataset.}
\label{tab:count-etiquetas-ara}
\end{table}

\setlength{\tabcolsep}{2pt} 
\renewcommand{\arraystretch}{1} 
\begin{table}[t!]
\centering 
\scriptsize
\begin{tabular}{l|c|c|c|c}
\multicolumn{ 1}{c|}{\multirow{2}{*}{Method}} & \multicolumn{ 1}{c|}{\multirow{2}{*}{Recall}} &  Relative & Recall without & Relative \\ 
 &  & Error     & popular GO labels     & Error \\
\hline
Classical [24] & 
\multicolumn{ 1}{c|}{\multirow{1}{*}{0.033}} & \multicolumn{ 1}{c|}{\multirow{1}{*}{-53.52\%}} &
\multicolumn{ 1}{c|}{\multirow{1}{*}{0.034}} & \multicolumn{ 1}{|c}{\multirow{1}{*}{-27.66\%}}\\ 
\hline 
Complete information &
\multicolumn{ 1}{c|}{\multirow{1}{*}{0.071}} & \multicolumn{ 1}{c|}{\multirow{1}{*}{-}} &
\multicolumn{ 1}{c|}{\multirow{1}{*}{0.047}} & \multicolumn{ 1}{|c}{\multirow{1}{*}{-}}\\ 
\hline
\gss{} method & 0.052 & -26.76\% & 0.038 & -19.15\% \\ 
\end{tabular} 
\vspace{\tfill{}}
\caption{Recall for the ARA dataset.}
\label{tab:accuracy_ara}
\vspace{\tffill{}}
\end{table}

\dif{Clustering performance was also evaluated for this dataset, comparing the clustering in Step 2 to the SOM-based co-clustering in terms of $BHI$ and $BC$. Results are shown in Table \ref{tab:results-comparison-ara}. It can be seen that the clustering quality of \gss{} outperforms the SOM co-clustering for this dataset as well. Our method yields the best $BHI$ (0.12) compared to the lower values (0.06 and 0.10) for the SOM-based method. Similar results are reached in terms of $BC$, with the best value (0.59) against the less compact results for the SOM approach.}

\begin{table}[t!]
\centering
\begin{tabular}{l|c|c}
\multicolumn{ 1}{c|}{Method} & BHI & BC \\ 
\hline
SOM [39], $m=0$ & 0.06 & 0.74 \\ 
\hline
SOM [39], $m=1$ & 0.10 & 0.82 \\ 
\hline 
\gss{} method & 0.12 & 0.59 \\ 
\end{tabular} 
\vspace{\tfill{}}
\caption{Comparison on clustering quality only for Step 2, ARA dataset.}
\label{tab:results-comparison-ara}
\vspace{\tftfill{}}
\end{table}

\begin{sidewaysfigure*}
\centerline{\includegraphics[width=\textwidth]{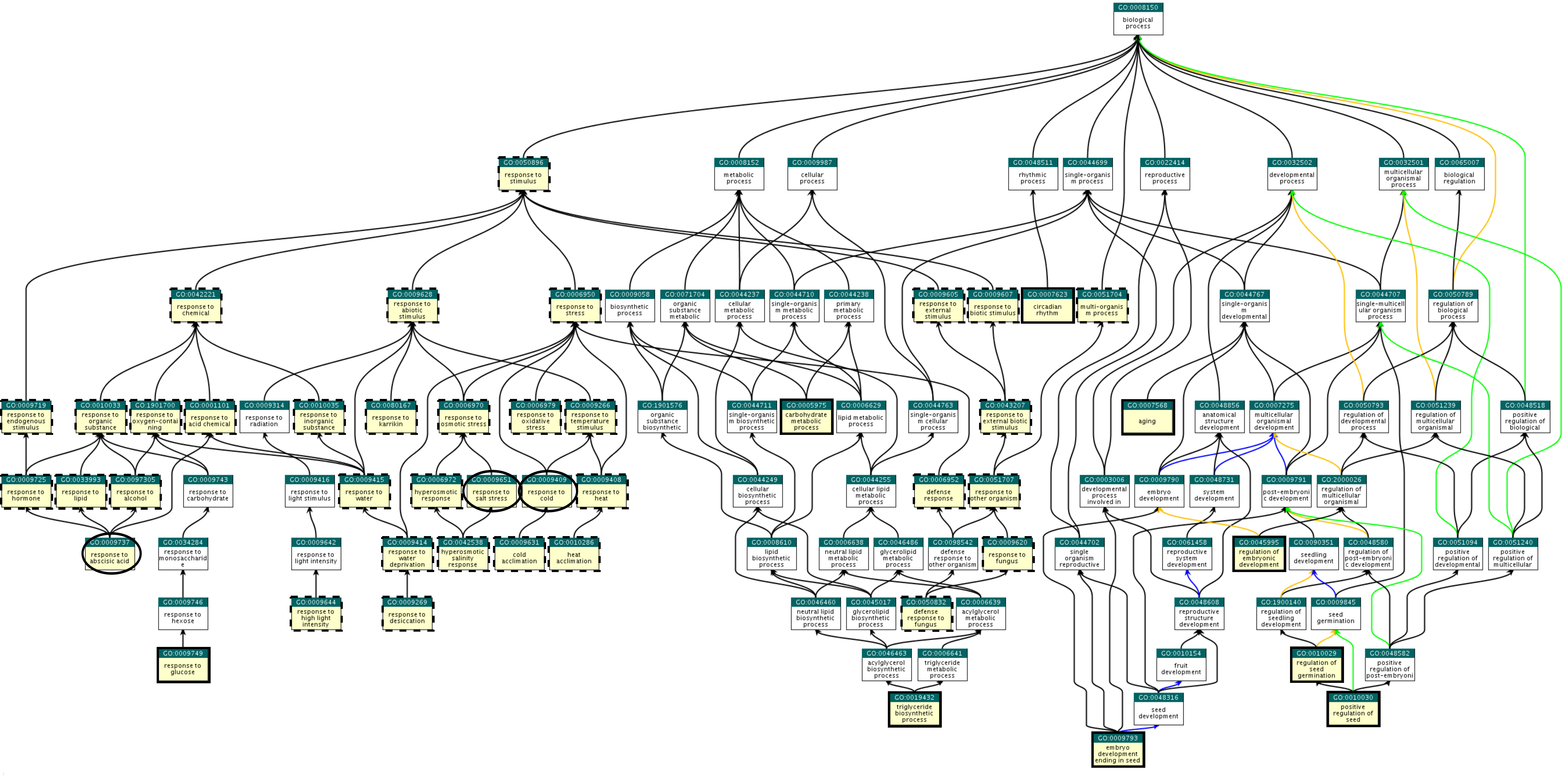}}
\caption{Inferred terms in the Gene Ontology for gene \geneexampleara{} from cluster \clusterara{} of the ARA dataset: assigned terms (dashed line), matching terms (ellipse), original B terms (thick line).} \label{fig:mapa_go_ara}
\end{sidewaysfigure*}

A detailed example regarding the automatic annotation of genes is also presented for this dataset. 
Table \ref{tab:enrichment-analysis-ara} shows the output for the enrichment analysis procedure. Results for cluster \clusterara{} are shown in table \ref{tab:validation-assignment-procedure-ara}, with matching terms shown in bold. Results are very promising for this example. Several actual GO terms for a subset of assigned genes matched exactly cluster terms obtained through enrichment analysis. This is shown graphically in Figure \ref{fig:mapa_go_ara}. Relative location of inferred terms in GO is shown in the figure, analogously to the example for the YEAST dataset. Matching terms are very close to the rest of the assigned terms. Original terms from $B$ are also close (as siblings) to assigned terms. From these results, it can be seen that there are many terms from $B$ that are correctly inferred by our approach. Furthermore, also for this dataset, it can be clearly seen that, for original terms from $B$ there is a remarkable closeness between them and the terms assigned by the procedure.

\begin{table}[t!]
\centering
\scriptsize
\begin{tabular}{r|c|l}
\textbf{$p$-value} & \textbf{GO term} & \textbf{Description} \\ \hline
 4.72e-18 & GO:0006950 & response to stress \\
 8.48e-17 & GO:0009628 & response to abiotic stimulus \\
 1.78e-16 & GO:0009409 & response to cold \\
 1.95e-16 & GO:0009266 & response to temperature stimulus \\
 1.82e-14 & GO:0050896 & response to stimulus \\
 9.3e-10 & GO:0009415 & response to water \\
 9.3e-10 & GO:0009414 & response to water deprivation \\
 3.71e-07 & GO:0006970 & response to osmotic stress \\
 8.74e-07 & GO:0080167 & response to karrikin \\
 1.83e-06 & GO:0009620 & response to fungus \\
 2.32e-06 & GO:1901700 & response to oxygen-containing compound \\
 8.15e-06 & GO:0009651 & response to salt stress \\
 1.02e-05 & GO:0009408 & response to heat \\
 2.68e-05 & GO:0009631 & cold acclimation \\
 5.25e-05 & GO:0050832 & defense response to fungus \\
 6.24e-05 & GO:0010035 & response to inorganic substance \\
 0.00016 & GO:0042221 & response to chemical \\
 0.000233 & GO:0009269 & response to desiccation \\
 0.000511 & GO:0006952 & defense response \\
 0.000581 & GO:0001101 & response to acid chemical \\
\end{tabular}
\vspace{\tfill{}}
\caption{Output of enrichment analysis for cluster \clusterara{} for ARA dataset.}
\label{tab:enrichment-analysis-ara}
\end{table}

\begin{table}[t!]
\centering
\scriptsize
\begin{tabular}{l|lll}
\multicolumn{1}{c|}{\textbf{Gene}} & \multicolumn{2}{c}{\textbf{GO actual terms}} \\ \hline
AT2G19450
 & \textbf{GO:0009409}  & GO:0010030 \\
 & \textbf{GO:0009651}  & GO:0019432 \\
 & \textbf{GO:0009737}  & GO:0045995 \\
 & GO:0005975 & GO:0007568 \\
 & GO:0007623 & GO:0009749 \\
 & GO:0009793 & GO:0010029 \\

AT1G64890
 & \textbf{GO:0009409}  & GO:0006810 \\
 & \textbf{GO:0009414}  & GO:0007623 \\
 & \textbf{GO:0009737} & \\

AT4G37470
 & \textbf{GO:0080167} & GO:0009640 \\
 & GO:0009704  & GO:0009744 \\
 & GO:0009813  & GO:0010224 \\

AT4G39260
 & \textbf{GO:0009409}  & GO:0000380 \\
 & \textbf{GO:0009651}  & GO:0007623 \\
 & \textbf{GO:0009737}  & GO:0010043 \\
 & GO:0045087 & \\

AT1G79440
 & \textbf{GO:0009408} & GO:0008152 \\
 & GO:0006333  & GO:0055114 \\
 & GO:0006540  & GO:0072593 \\
 & GO:0009416  & GO:0009450 \\

\end{tabular}
\vspace{\tfill{}}
\caption{Validation results for some genes from B assigned to cluster \clusterara{} for ARA dataset. Original matching labels shown in bold}
\label{tab:validation-assignment-procedure-ara}
\end{table}

\end{document}